\documentclass[10pt,twocolumn,letterpaper]{article}

\usepackage[pagenumbers]{cvpr} 

\usepackage{graphicx}
\usepackage{amsmath}
\usepackage{amssymb}
\usepackage{booktabs}

\usepackage[pagebackref,breaklinks,colorlinks]{hyperref}

\usepackage[capitalize]{cleveref}
\crefname{section}{Sec.}{Secs.}
\Crefname{section}{Section}{Sections}
\Crefname{table}{Table}{Tables}
\crefname{table}{Tab.}{Tabs.}

\makeatletter
\@namedef{ver@everyshi.sty}{}
\makeatother

\usepackage{amsmath}
\usepackage{amsfonts}
\usepackage{amssymb}
\usepackage{wrapfig}
\usepackage{subcaption}
\usepackage{multirow}
 \usepackage{mathtools} 

\usepackage{verbatim}

\usepackage{anyfontsize}

\usepackage{microtype}
\usepackage{graphicx}
\usepackage{booktabs} 
\usepackage{multirow}
\usepackage{amsmath,amssymb}
\usepackage{booktabs}
\usepackage{caption,subcaption}

\usepackage{xcolor}
\definecolor{mygreen}{HTML}{3cb44b}
\definecolor{skyblue}{HTML}{beffff}
\definecolor{lightgreen}{HTML}{90ee90}

\usepackage{tcolorbox}
\usepackage{enumitem}
\setitemize{itemsep=10pt,topsep=0pt,parsep=0pt,partopsep=0pt}
\usepackage{adjustbox}

\newcommand{\RN}[1]{%
	\textup{\lowercase\expandafter{\it \romannumeral#1}}%
}
\usepackage{tabu}

\usepackage{amsmath}
\DeclareMathOperator*{\argmax}{argmax}

\newcommand{\beq}{\vspace{0mm}\begin{equation}}
\newcommand{\eeq}{\vspace{0mm}\end{equation}}
\newcommand{\beqs}{\vspace{0mm}\begin{eqnarray}}
\newcommand{\eeqs}{\vspace{0mm}\end{eqnarray}}
\newcommand{\barr}{\begin{array}}
\newcommand{\earr}{\end{array}}

\newcommand{\qv}[0]{{\boldsymbol{q}}}

\newcommand{\sv}{{\boldsymbol{s}}}
\newcommand{\tv}[0]{{\boldsymbol{t}}}
\newcommand{\uv}{\boldsymbol{u}}
\newcommand{\vv}{\boldsymbol{v}}

\newcommand{\xv}{\boldsymbol{x}}

\newcommand{\thetav}{\boldsymbol{\theta}}

\newcommand{\phiv}{\boldsymbol{\phi}}

\newcommand{\R}{\mathbb{R}}

\newcommand{\Xcal}{\mathcal{X}}
\newcommand{\Ycal}{\mathcal{Y}}

\newcommand{\Bcal}{\mathcal{B}}

\newcommand{\Tcal}{\mathcal{T}}

\newcommand{\Lcal}{\mathcal{L}}

\newcommand{\Pcal}{\mathcal{P}}
\newcommand{\Ical}{\mathcal{I}}

\newcommand{\Qcal}{\mathcal{Q}}

\newcommand{\Scal}{\mathcal{S}}

\usepackage{color, colortbl}
\definecolor{Gray}{gray}{0.93}

\usepackage{lipsum}

\newcommand\blfootnote[1]{%
  \begingroup
  \renewcommand\thefootnote{}\footnote{#1}%
  \addtocounter{footnote}{-1}%
  \endgroup
}

\usepackage{xcolor,amsmath}
\usepackage[linesnumbered,ruled,vlined]{algorithm2e}
\DontPrintSemicolon

\usepackage{color, colortbl}

\definecolor{emerald}{rgb}{0.31, 0.78, 0.37}
\definecolor{coralred}{rgb}{1.0, 0.25, 0.25}

\newcommand{\MyColorBox}[2][red]%
{%
    \settowidth{\Width}{#2}%
    \colorbox{#1}%
    {%
        \raisebox{-\DepthReference}%
        {%
                \parbox[b][\HeightReference+\DepthReference][c]{\Width}{\centering#2}%
        }%
    }%
}

\definecolor{codegray}{gray}{0.9}

\SetKwComment{Comment}{\color{green!50!black}\# }{}

\SetKwProg{Function}{def}{:}{}

\SetKwProg{For}{for}{:}{}
\SetKwProg{If}{if}{:}{}

\def\cvprPaperID{***} 
\def\confName{CVPR}
\def\confYear{2023}

\newlength\savewidth\newcommand\shline{\noalign{\global\savewidth\arrayrulewidth
  \global\arrayrulewidth 1pt}\hline\noalign{\global\arrayrulewidth\savewidth}}

\renewcommand{\paragraph}[1]{\vspace{1.25mm}\noindent\textbf{#1}}

\usepackage{xcolor}

\usepackage{pifont}
\newcommand{\cmark}{\text{\ding{51}}}
\newcommand{\xmark}{\text{\ding{55}}}

\newcommand{\shortname}{\textsc{React}}
\newcommand{\longname}{\textbf{RE}trieval-\textbf{A}ugmented \textbf{C}us\textbf{T}omization}

\begin{document}

\title{Learning Customized Visual Models with Retrieval-Augmented Knowledge}

\author{Haotian Liu \quad Kilho Son \quad Jianwei Yang \quad Ce Liu \quad Jianfeng Gao \quad Yong Jae Lee\textsuperscript{$\P\ddagger$} \quad Chunyuan Li\textsuperscript{$\P\ddagger$}
}

\author{
\normalsize{
    Haotian Liu$^{\dagger\S\spadesuit}$ \quad Kilho Son$^{\ddagger}$ \quad Jianwei Yang$^{\ddagger}$ \quad Ce Liu$^{\ddagger}$ \quad Jianfeng Gao$^{\ddagger}$ \quad Yong Jae Lee\textsuperscript{$\dagger\P$} \quad Chunyuan Li\textsuperscript{$\ddagger\P\spadesuit$}
}
\and
{
\normalsize
$^{\dagger}$ \textbf{University of Wisconsin--Madison}
\quad\quad\quad
$^{\ddagger}$ \textbf{Microsoft} \;
}
\and
\centerline{\tt\footnotesize  
\{lht,yongjaelee\}@cs.wisc.edu 
 \ \{kilhoson,jianwyan,ce.liu,jfgao,chunyl\}@microsoft.com
}
\and
\centerline{\normalsize \url{https://react-vl.github.io}
}
}
\maketitle

\begin{abstract}

Image-text contrastive learning models such as CLIP have demonstrated strong task transfer ability. 
The high generality and usability of these visual models  is achieved via a web-scale data collection process to ensure broad concept coverage, followed by expensive pre-training to feed all the knowledge into model weights. 
Alternatively, we propose \textbf{\shortname{}}, \longname{}, a framework to acquire the relevant web knowledge to build customized visual models for target domains.
We retrieve the most relevant image-text pairs ($\sim$3\% of CLIP pre-training data) from the web-scale database as external knowledge, and propose to customize the model by only training new modualized blocks while freezing all the original weights.
The effectiveness of \shortname{} is demonstrated via extensive experiments on classification, retrieval, detection and segmentation tasks, including zero, few, and full-shot settings.
Particularly, on the zero-shot classification task, compared with CLIP, it achieves up to 5.4\% improvement on ImageNet and 3.7\% on the \textsc{Elevater} benchmark (20 datasets).

\end{abstract}

\blfootnote{${\spadesuit}$~core contribution; ${\P}$~equal advising; $\S$ work initiated during an internship at Microsoft.}

\section{Introduction}
\label{sec:intro}

It has been a fundamental research problem in computer vision (CV) to
build a transferable visual system that can easily adapt to a wide range of downstream tasks. With remarkable advances in deep learning, a de facto solution to achieve this is to train deep neural networks on a large amount of data to pursue the so-called {\it generic} visual representations. This dates back to the standard supervised training on ImageNet~\cite{deng2009imagenet}, whose superb representation power is further demonstrated in BiT~\cite{kolesnikov2020big}/ViT~\cite{dosovitskiy2020image} by scaling up the training to JFT300M~\cite{sun2017revisiting}. Along the way, recent efforts have been applied to the popular image self-supervised learning~\cite{he2020momentum,chen2020simple,he2022masked} to reduce the demand for labeled data. The third approach is image-text contrastive learning trained on billion-scale web-crawled image-text pairs. Such models, like CLIP~\cite{radford2021learning} and ALIGN~\cite{jia2021scaling}, are able to achieve great performance on different downstream domains, without the need of any human labels.

Excellent empirical performance has been achieved with the above three pre-training methods, by following the well established two-stage {\it pre-training then adaptation} pipeline: model pre-training from scratch on large data, then model adaptation directly on downstream tasks. Specifically, the pre-trained models are adapted to downstream tasks by considering the available task-specific samples only: either evaluated in a zero-shot task transfer manner, or updated using linear probing (LP)~\cite{radford2021learning}, finetuning (FT)~\cite{li2022elevater}, or prompt tuning~\cite{zhou2022learning,saito2022prefix}. 
Following this two-stage pipeline, most research has reverted to the faith that building transferable visual systems is equivalent to developing more generic visual models by feeding all knowledge in the model pre-training stage. Therefore, the community has been witnessing a trend in exploring scaling success of pre-training model and data size with less care on the target domain,  hoping that the model can adapt to any downstream scenario. 

\begin{figure*}[t!]
	\vspace{-0mm}
	\centering
\includegraphics[height=4.3cm]{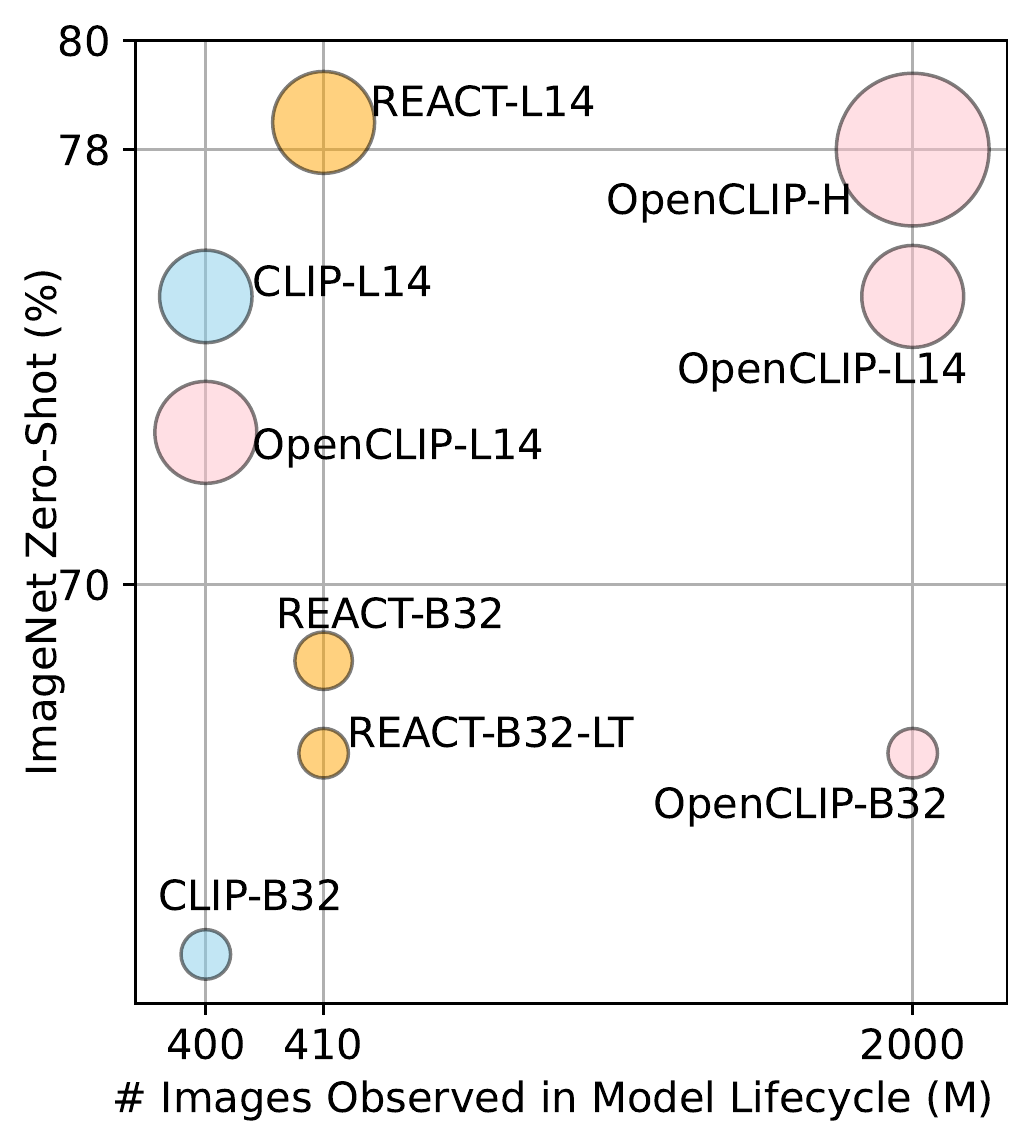}
\includegraphics[height=4.3cm]{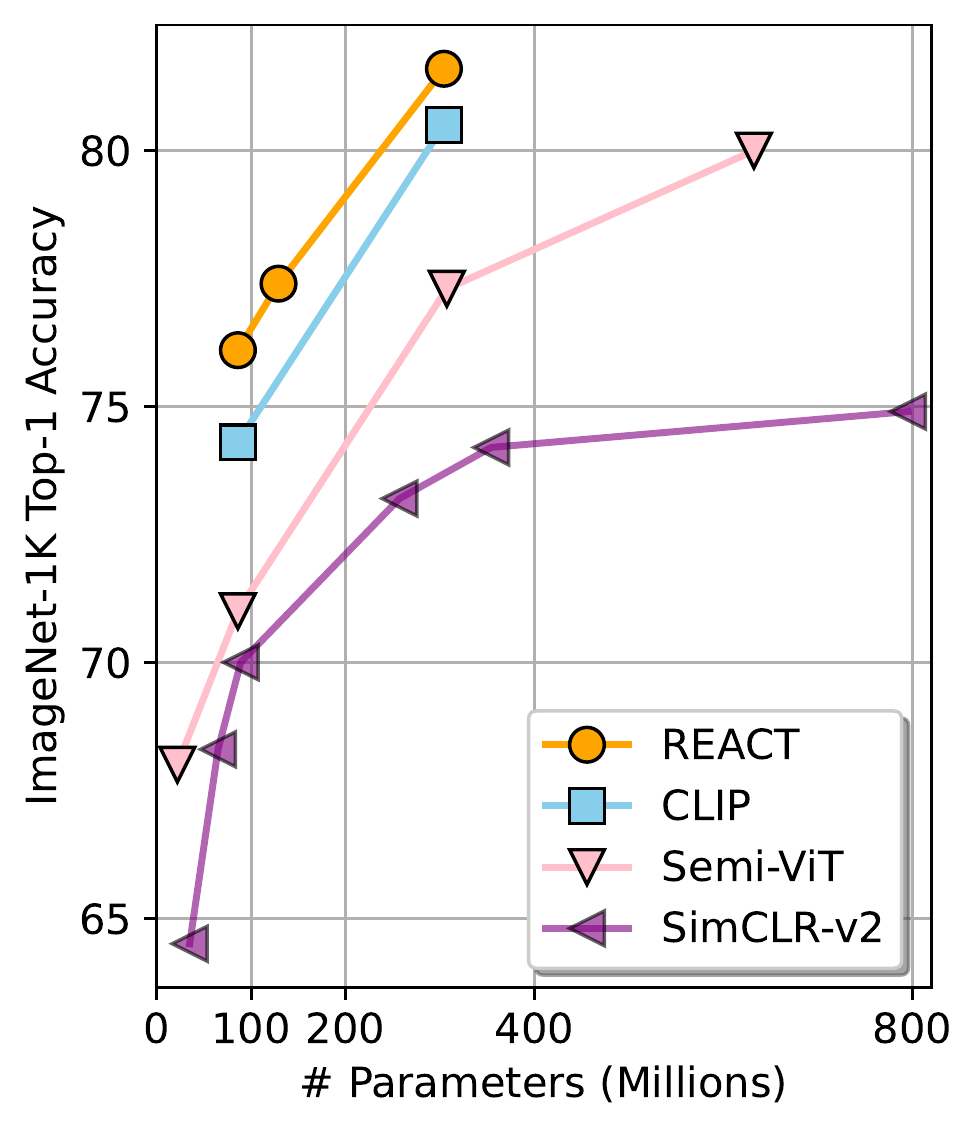} ~
\includegraphics[height=4.3cm]{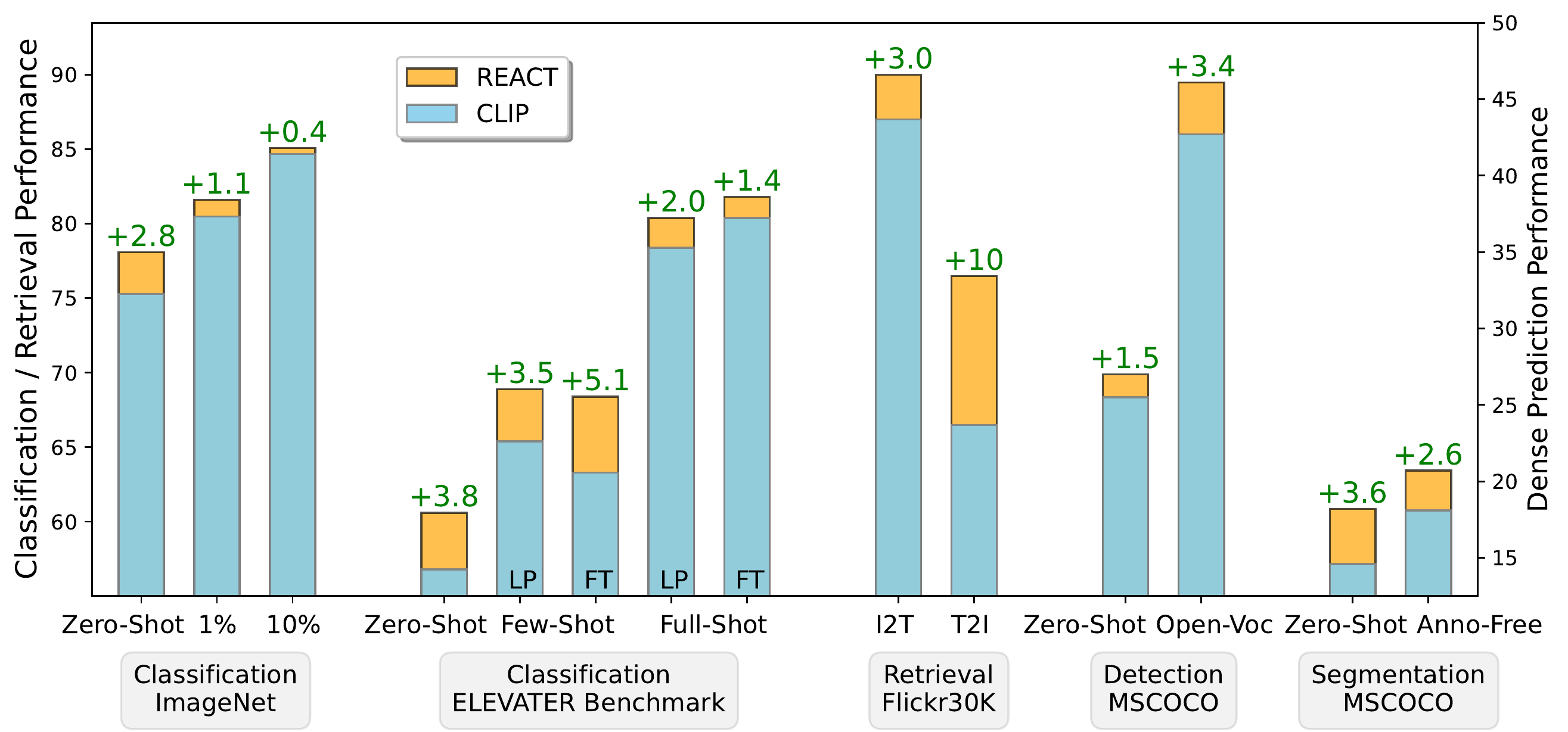} ~
	\vspace{-2mm}
	\caption{\shortname{} achieves the best zero-shot ImageNet performance among public checkpoints  with nearly $5\times$ smaller data size (Left), achieves new SoTA on semi-supervised ImageNet classification in the 1\% labelled data setting (Middle), and consistently transfer better than CLIP on across a variety of tasks, including ImageNet classification,  zero/few/full-shot classification on 20 datasets in ELEVATER benchmark, image-text retrieval, object detection and segmentation (Right). Please see the detailed numbers and settings in the experimental section. For the left figure, circle size indicates model size.}
	\vspace{-4mm}
	\label{fig:teaser}
\end{figure*}

In this paper, we argue that the conventional two-stage pipeline above is over-simplified and less efficient, in achieving the goal of building a transferable visual system in real-world settings. Instead, we propose a {\it customization} stage in between the pre-training and adaptation, where customization is implemented  by systematically leveraging retrieved external knowledge. The inspiration comes from how humans are specialized in society for better generalization: instead of trying to memorize all concepts, humans are trained/prepared in a relevant subject to master a certain skill, while maintaining the basic skills in pre-training. 

To this end, we explore a systematic approach to acquire and learn with external knowledge sources
from a large image-text corpus for model customization. The process of collecting external image-text knowledge is fully automatic without extra human annotation. The acquired knowledge typically contains richer information about the concept: relevant images that never appear in the downstream training and evaluation set, and richer text descriptions about concept semantics. Such multi-modal knowledge sources are generally available on the web, and further open-sourced like LAION~\cite{schuhmann2021laion,schuhmann2022laion}. They cover a variety of
domains, making it possible to develop customized visual models for task-level transfer. Similar retrieval-augmented intuitions have been exploited in computer vision for class-level transfer~\cite{long2022retrieval}, but not yet for task-level transfer (similar to that of CLIP).
Our main findings/contributions can be summarized as follows.

{\it We propose to explore the potential of the web-scale image-text corpus as external knowledge} to significantly improve task-level transfer performance on the target domain at an affordable cost.
A simple and effective strategy is proposed. To begin with, we build a large-scale multi-modal indexing system to retrieve the relevant image-text pairs using CLIP features and approximate nearest neighbor search. For a CV problem, the task instruction is often sufficiently specified with text such as class names, which allows us to utilize them as queries to retrieve the relevant image-text pair knowledge from the indexing system. \emph{No images from the CV problem are needed.} To efficiently build the customized visual model, we propose a novel modularized learning strategy: only updating the additional trainable weights on the retrieved knowledge, and freezing the original model weights. Hence, the model masters the new skill without forgetting basic skills.

{\it The generality and effectiveness of the proposed customization strategy is demonstrated on four CV problems}.
We instantiate it with CLIP, and develop the customized visual models for image classification on ImageNet and 20 datasets in \textsc{Elevater}~\cite{li2022elevater}, image-text retrieval on COCO~\cite{lin2014microsoft}/Flickr~\cite{plummer2015flickr30k}, as well as object detection and semantic segmentation on COCO~\cite{lin2014microsoft}.
The knowledge bases are considered as LAION~\cite{schuhmann2021laion} and larger web-crawled multi-modal data. The retrieval-augmented knowledge ($\sim$3\% image-text pairs compared with the original training data) significantly improves the model's zero-shot performance without the need of accessing any images on downstream tasks. See Figure~\ref{fig:teaser} for highlighted results.
For example, our ViT-L/14 checkpoint achieves 78.5\% zero-shot accuracy on ImageNet~\cite{deng2009imagenet}, surpassing all public checkpoints from CLIP~\cite{radford2021learning} and OpenCLIP~\cite{openclip}, including those with larger model size and trained on a much larger LAION-2B~\cite{schuhmann2022laion}. The new customized visual models also demonstrate higher few/full-shot performance than the original generic model counterparts. 

{\it Our retrieval system, codebase, and pre-trained models will be publicly available}.  To make this line of research more accessible, our retrieved subsets for both \textsc{Elevater} and ImageNet will also be made available, with an easy-to-use toolkit to download the subsets without storing the whole dataset locally.  It poses a feasible direction for leveraging the ever-increasing data from the Internet for customized visual recognition, especially for the low-resource regimes. 

\vspace{-2mm}
\section{Related Work}
\label{sec:related}
\vspace{-2mm}
\paragraph{Vision-Language Pretraining.}
Learning transferable visual representations from natural language supervision is an emerging research area.  The pioneering works of CLIP~\cite{radford2021learning} and ALIGN~\cite{jia2021scaling} make use of contrastive learning to pretrain models on billion-scale web-crawled image-text pairs. There are an increasing number of studies to improve their generality from various modeling perspectives, including training objectives~\cite{geng2022multimodal,dong2022maskclip,yu2022coca,mustafa2022multimodal,zhai2022lit,gao2022pyramidclip}, scaling techniques~\cite{chen2022pali,yu2022coca,pham2021combined}, data efficiency~\cite{li2021supervision,lee2022uniclip}, and leveraging multilingual correlations~\cite{jain2021mural,chen2022pali}.
In academia, several works demonstrate techniques to improve the learned semantic representations on datasets at a smaller scale (\eg CC3M~\cite{sharma2018conceptual}, CC12M~\cite{changpinyo2021conceptual}, YFCC15M~\cite{radford2021learning,thomee2016yfcc100m}), by exploring pretraining on a unified image-text-label space~\cite{yang2022unicl}, token-level contrastive loss~\cite{yao2021filip}, and auxiliary within-modality contrastive loss~\cite{mu2021slip,yang2022vision,you2022learning}.
Complementary to the above works, we build on top of existing pre-trained generic models, and aim to improve the model's performance by customizing them using retrieved \emph{relevant} image-text pairs.

\paragraph{Retrieval-Augmented Models.}
In natural language processing, several works augment large language models with external data encoded with structured language and relation representations~\cite{peters2019knowledge,guu2020realm,lewis2020retrieval,liu2020k,yu2021dict,borgeaud2021improving,khandelwal2019generalization}.
Motivated by retrieval-augmented models in NLP, several recent works leverage visual and / or textual knowledge to improve classification~\cite{long2022retrieval}, question answering~\cite{wu2021multi,marino2021krisp,yang2021empirical,chen2022murag}, image generation~\cite{blattmann2022retrieval,sheynin2022knn,chen2022re,zhou2022lafite2}, and multi-modal tasks simultaneously ~\cite{yasunaga2022retrieval}.  RAC~\cite{long2022retrieval} improves long-tail classification by retrieving from a non-parametric memory consisting of pre-encoded images and text. K-LITE~\cite{shen2022klite} enhances the text prompts with the retrieved external knowledge that is encoded in natural language.
Our paper leverages the paired knowledge of image-text and aims to improve task transfer performance for core vision problems such as classification, retrieval, detection and segmentation.

\paragraph{Adaptation of Vision-Language models.}
CLIP demonstrates impressive zero-shot and linear probing performance on different downstream domains.  Several works explore improving the domain adaptation performance on CLIP models.  \textsc{Elevater}~\cite{li2022elevater} leverages the text encoder outputs to initialize the task-specific linear head to improve the linear probe and finetuning performance of CLIP.  Inspired by prompting techniques in NLP, recent works~\cite{zhou2022learning,saito2022prefix} make use of learnable prompts that are trained on a few samples on downstream tasks.
Similar to these works, this paper aims to improve CLIP's performance on downstream tasks, while making use of relevant image-text pairs data to improve the model's performance, without access to the downstream images.  Furthermore, when downstream samples are available, they are complimentary to our method.

\vspace{-2mm}
\section{Retrieval-Augmented Customization}
\label{sec:approach}
\vspace{-1mm}

\subsection{Preliminaries}
\label{sec:preliminaries}
\vspace{-1mm}
 Computer vision models have achieved strong transfer performance, when learning with large-scale image data only~\cite{he2020momentum}, image-label data~\cite{kolesnikov2020big} and/or image-caption data~\cite{radford2021learning,yang2022unicl,yuan2021florence}. Without loss of generality, we follow~\cite{yang2022unicl} and define a unified triplet-wise format $(\xv, \tv, y)$ for image-text-label data, where $\xv \in \Xcal$  is an image, $\tv \in \Tcal$ is its  language description, and $y \in \Ycal$ is a label indicating the index of the unique language description in the dataset. In a general form, the language description is a text sequence $\tv = [t_1, \cdots, t_L]$. It ranges from simple category names representing visual concepts when $L$ is small, to more free-form and  semantic-rich sentences such as captions when $L$ is relatively large. 
 
A typical transfer learning pipeline follows the procedure of {\it pre-training then adaptation}:
$(i)$ 
With large-scale pre-training, an image encoder foundation model $f_{\thetav}$ parameterized by $\thetav$ is first trained to represent image $\xv$ as a visual feature vector $ \Tilde{\vv}  \in \R^{P\times 1}$: $ \Tilde{\vv} = f_{\thetav}(\xv)$. For recent language-image models~\cite{radford2021learning}, a dual-encoder architecture is often employed, where an additional text encoder $f_{\phiv}(\tv)$ parameterized by $\phiv$ represents the sentence $ \Tilde{\uv}  \in \R^{P \times 1}: \Tilde{\uv}  = f_{\phiv}(\tv)$.
$(ii)$ 
Given a downstream task, model adaptation is typically performed using the available task-specific information, or {\it task instruction} $\Ical$. For example, the task-level transfer of a language-image model is described as:

\begin{itemize}[leftmargin=4.5mm]
\vspace{1mm}
\item {\it Zero-shot.} In a customized setting, the simplest task definition can be provided as a set of category names for visual recognition, leading to the task instruction $\Ical_0 = \{\tv\}$. No training image $\xv$ is available, not to mention the corresponding label $y$. 

\vspace{-2mm}
\item {\it Few/Full-shot.} The users may spend annotation cost to curate $N$ image-label pairs as the training instances, making the task instruction more specific,  $\Ical_{\text{F}}  = \{ (\xv_n, \tv_n, y_n) \}_{n=1}^N$, which allows updating the image encoder model $f_{\thetav}$ for better adaptation performance.

\vspace{1mm}
\end{itemize}

\begin{figure}[t]
	\centering
	\includegraphics[width=0.95\linewidth]{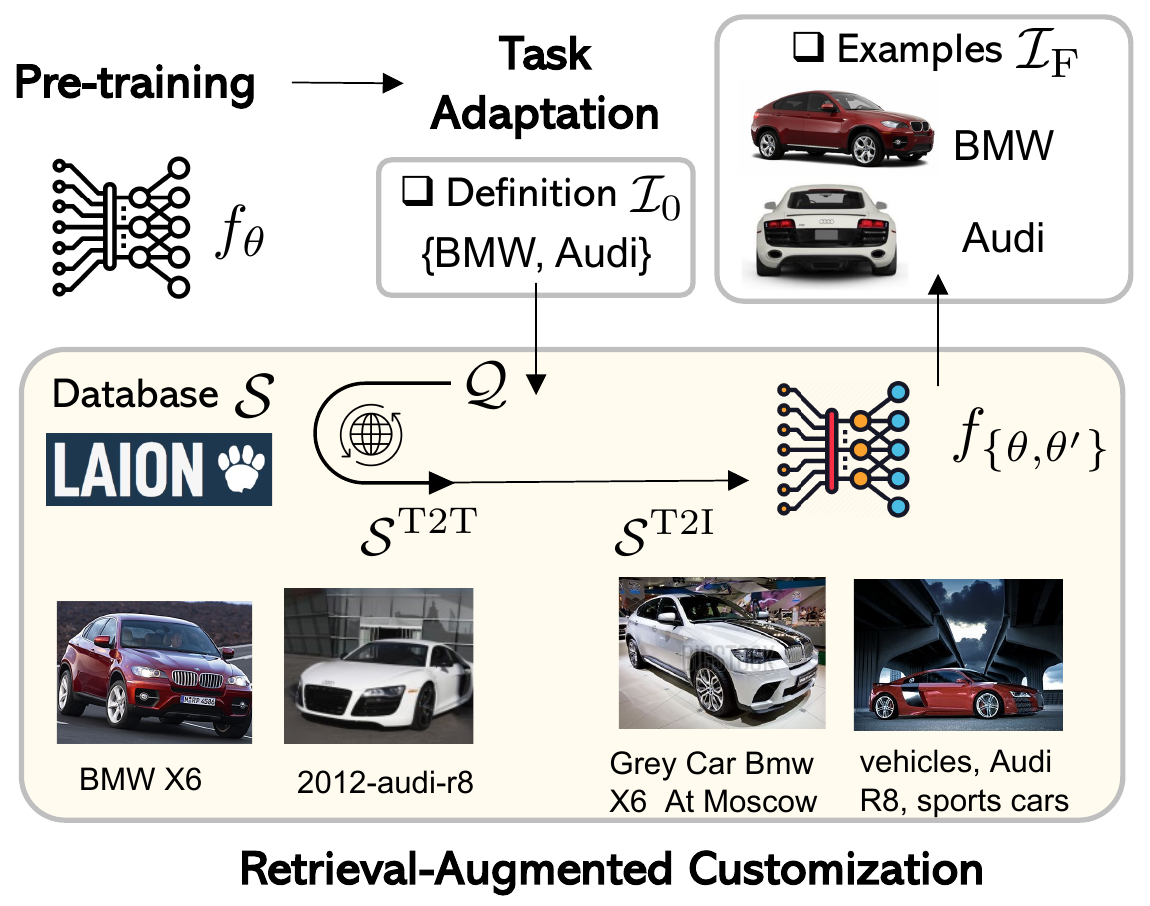}
    \vspace{-2mm}
    \caption{Illustration of the proposed \shortname{} framework.}
    \label{fig:react}
    \vspace{-4mm}
\end{figure}

In this paper, we assume there exists a web-scale image-text corpus as the external knowledge source $\Scal  = \{ (\xv_m, \tv_m) \}_{m=1}^M$, where $M$ is the database size, \eg 400M for LAION~\cite{schuhmann2021laion}. One may use the task instruction $\Ical$ as a query to seek additional relevant knowledge to build a more transferable visual system.
Given the downstream task instruction $\Ical$ and an external knowledge source $\Scal$, our goal is to learn customized visual-semantic representations, which are readily transferable to the downstream task of interest, whose training and evaluation images are not observed during the customization process. To this end, we propose \shortname{}. We illustrate the high-level idea in Figure~\ref{fig:react}, and describe the process as follows.

\subsection{Multi-modal External Knowledge}
\vspace{-2mm}
\paragraph{Knowledge Base Construction.}
We explore web-scale image-text data as the multi-modal knowledge base $\Scal$ in this paper. Ideally, one may consider the entire web as the  knowledge base, and use Google or Bing search to retrieve the relevant knowledge.
We consider two large static datasets with image-text pairs.  To control the experiment complexity and ensure reproducibility, we use LAION-400M~\cite{schuhmann2021laion}, a publicly available database with 400M pairs, for most of the experiments.  To further study the scaling influence of the retrieval base, we conduct comparisons on Web-800M, a privately collected web database with 800M pairs.

To facilitate an efficient knowledge acquisition process, we use pre-trained contrastive models (\eg CLIP) as the feature extractor, and build a cross-modal retrieval system using FAISS~\cite{johnson2019faiss}. We use its Hierarchical Navigable Small World (HNSW) approximate $k$-NN lookup~\cite{malkov2018hnsw} to balance performance and efficiency.  For more details, please refer to the supplementary materials.
After the retrieval system is built on the designated retrieval pool, it can be efficiently used for retrieving relevant image-text pairs for \emph{various} downstream domains.

\begin{figure*}[ht!]
	\vspace{-0mm}\centering
\includegraphics[height=3.0cm]{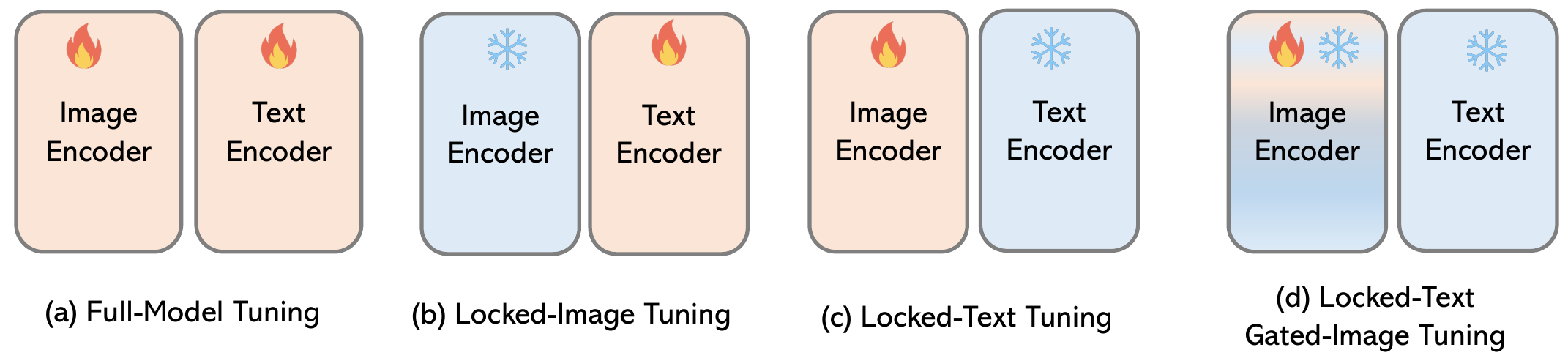} ~
\includegraphics[height=3.0cm]{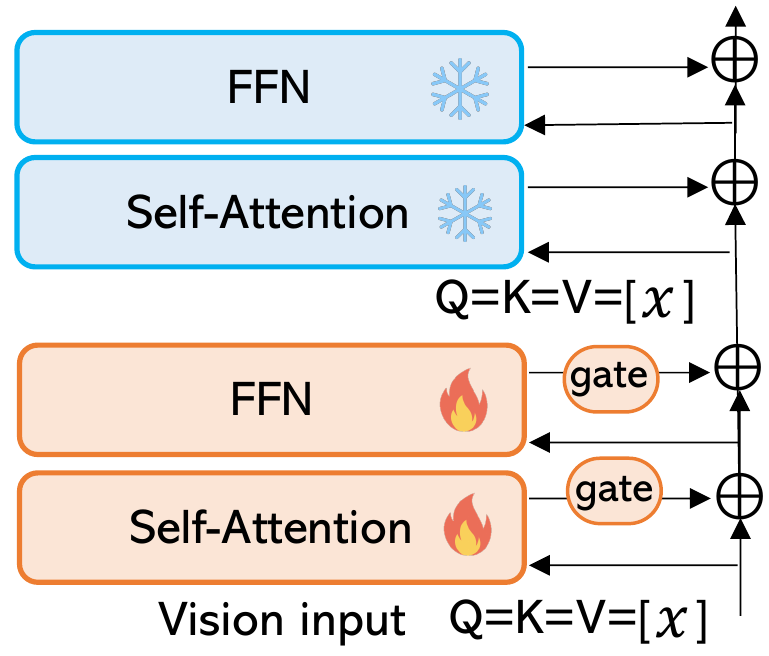} 
	\vspace{-2mm}
	\caption{Illustrative comparisons across different model tuning methods. (a) and (b) are existing baseline tuning methods. For model customization in a target domain, we found that (c) and (d) work better. One layer of the proposed  modularized image encoder in locked-text gated-image tuning is illustrated in right side.
	 }
	\vspace{-3mm}
	\label{fig:method_comparison}
\end{figure*}

\paragraph{Retrieval-Augmented Task Instruction}
To facilitate the same interface for various customized visual tasks in the wild, it is desirable to have the same uniform task instruction schema. 
In NLP,  all task instructions can follow the same uniform schema, composed of {\it task definition} and {\it positive/negative examples}~\cite{mishra2021cross,wang2022benchmarking}. Here, the task definition defines a given task in natural language, completely specifying how an input is expected to be mapped to an output text. We note a coherence connection between this NLP task schema and the customized  zero/few/full-shot CV settings in Section~\ref{sec:preliminaries}. Following a similar schema, the minimum requirement to specify a visual task is the task definition $\Ical_{\text{0}}$, where category names illustrate the target visual concepts in natural language. Though adding human-annotated examples is a natural way to clarify the task and yield the complete schema $\Ical_{\text{F}}$, extra cost is introduced. 

It is of high interest to clarify the task using relevant examples, without human curating cost. Therefore, we propose to augment the task instruction with the retrieved examples from the external multi-modal knowledge base $\Scal$. 
For each concept $\tv \in \Ical_{\text{0}}$ in a given task, we first represent it in natural language $\qv = g_{prompt}(\tv)$ using the language prompt as in \cite{radford2021learning}, through inserting the concept into a set of task-specific templates $\Pcal$. The task definition is expanded in its natural language form: 
\begin{align} \mathcal{Q} = \{\qv ~|~ \qv = g_{prompt}(\tv), \forall \tv \in \Ical_{\text{0}}, prompt \in \Pcal  \}. 
\end{align}

Next, we perform our knowledge retrieval process to acquire the relevant image-text pair $\sv = g_{retrieve}(\qv)$ from the source $\mathcal{S}$. Two types of retrieval processes are considered to acquire the top-$K$ pairs: 

\begin{itemize}[leftmargin=4.5mm]
\vspace{1mm}
\item  Text-to-Text (T2T) retrieval allows us to retrieve more relevant examples as they have a better match with our target concept. The T2T-retrieved set for $\Ical_{\text{0}}$ is:
\begin{align}\label{eq:t2t_retrieved_set}
\hspace{-3mm}
\small 
\Scal^{T2T} \!=\! \{  (\xv, \tv) \in \Scal: \! \underset{\tv \in \mathbb{T}, |\mathbb{T}| = K }{\argmax} f_{\phiv}(\tv)^{\top}\! f_{\phiv}(\qv),   \forall \qv \in \Qcal \}
\end{align}

\vspace{-3mm}
\item  
Text-to-image (T2I) retrieval allows us to have more diversity in the text descriptions in our retrieved examples. The T2I-retrieved set for $\Ical_{\text{0}}$ is:
\begin{align}\label{eq:t2i_retrieved_set}
 \hspace{-3mm}
\small 
\Scal^{T2I} \!=\! \{  (\xv, \tv) \in \Scal: \! \underset{\xv \in \mathbb{X}, |\mathbb{X}| = K }{\argmax} f_{\thetav}(\xv)^{\top}\! f_{\phiv}(\qv),   \forall \qv \in \Qcal\}
\end{align}
\end{itemize}

Both $\Scal^{T2T} $ and $\Scal^{T2I}$ are retrieved examples to augment the task definition $\Ical_{\text{0}}$, without accessing the images in the training or validation set of the task. Compared to $\Ical_{\text{F}}$, they are ``free'' external knowledge to clarify the task and can be used to build a more transferable system.

\subsection{Model Customization}
\vspace{-2mm}

After retrieving the relevant multi-modal examples, 
one may employ the naive customization solution by fine-tuning the full-model initialized from pre-trained weights, as in Figure~\ref{fig:method_comparison}(a).
Alternatively, we propose an affordable solution to endow pre-trained models with a new capability to leverage this external knowledge.
The pre-trained generic visual models have gained strong transfer
abilities and access to a large amount of internal knowledge stored in the model weights. We freeze the weights of these models so that their initial capacity remains unchanged. 
To bridge these pretrained models harmoniously to the customized domain, we consider {\it  locked-text gated-image tuning} with the following two techniques, illustrated in Figure~\ref{fig:method_comparison}(d). 

\paragraph{Modularized Image Encoder. } In order to provide sufficient expressivity to the model and make it able to adapt well on retrieved knowledge, we insert gated self-attention dense blocks in between the original layers of the image encoder, and train the new blocks from scratch. Those blocks are made of a self-attention layer, that attends the early layer inputs, followed by an extra dense feed-forward layer. Please see a visual illustration of this gated block in the rightmost of Figure~\ref{fig:method_comparison}(d).
We denote the parameters of all new modules as $\thetav'$. This design is inspired by the gated cross-attention-dense blocks in Flamingo~\cite{alayrac2022flamingo}  and frozen multi-modal model~\cite{tsimpoukelli2021multimodal}. The difference is that the trainable module is introduced in Flamingo to enable cross-modal conditioning, while we adapt it for model growing in new customized domains. 

\paragraph{Frozen Text Encoder.} The text encoder in language-image contrast models represents the task semantic space. To maintain it, we propose {\it locked-text tuning}, which freezes the text model weights so that the generic task encoding knowledge remains locked; see Figure~\ref{fig:method_comparison}(c).   This is in contrast with {\it locked-image tuning} (LiT)~\cite{zhai2022lit} in Figure~\ref{fig:method_comparison}(b), where the image encoder is frozen and the text encoder is fine-tuned, which teaches a text model to read out good representations from a pre-trained image model for new tasks.

We extract the normalized feature vectors in a hyper-sphere using $ \uv_i = \frac{    f_{ \{\thetav, \thetav'\}}(\xv_i)  }{   \| f_{\{\thetav, \thetav'\}}(\xv_i)  \|} $ and $ \vv_j = \frac{  f_{\phiv}(\tv_j)   }{ \| f_{\phiv}(\tv_j)\| }  $. To customize the model wrt task definition $\Ical_0$, we update $\thetav'$ using a bidirectional learning objective between images and language on the retrieved knowledge pool $\Scal^{T2T} \text{and/or}~ \Scal^{T2I}$:
\begin{align}\label{eq:obj_unicl}
&	\min_{ \{ \thetav' \} } ~~ \Lcal_{\text{C}} 	=  \Lcal_{i2t} + \Lcal_{t2i}, ~\text{with}~ 
\Bcal \sim \Scal^{T2T} \text{or}~ \Scal^{T2I}\\
	\hspace{-2mm}
& 
\small{ 
\Lcal_{i2t}\! =\! - \sum_{ i \in \Bcal } \frac{1}{ |\Pcal(i)|  }\!  \sum_{ k \in \Pcal(i) }
\!\log \frac{ \exp(\tau \uv_{i}^{\top} \vv_k)  }{\sum_{ j \in \Bcal}  \exp(\tau \uv_{i}^{\top} \vv_{j})  }
} ~~~\text{and}~ \\
 & 
 \small{ 
\Lcal_{t2i}\!	= \! - \sum_{ j \in \Bcal } \frac{1}{ |\Qcal(j)|  }\!  \sum_{ k \in \Qcal(j) }
\!\log \frac{ \exp(\tau \uv_{k}^{\top} \vv_j )  }{\sum_{ i \in \Bcal}  \exp(\tau \uv_{i}^{\top} \vv_{j} )  }
}
\end{align}
where $\tau$ is a temperature hyper-parameter controlling the strength of penalties on hard negative samples, and $  \Pcal(i) = \{ k | k \in \Bcal, \vv_k^{\top} \vv_i \ge\gamma\}$, $  \Qcal(j) = \{ k | k \in \Bcal, \vv_k^{\top} \vv_j \ge\gamma\} $. We set $\gamma=0.9$ for classification tasks to force image-text pairs sharing the similar text to be positive. Note \eqref{eq:obj_unicl} is a general form; it  reduces to UniCL~\cite{yang2022unicl}  when $\gamma=1.0$; it further reduces to the training objective of CLIP~\cite{radford2021learning} or ALIGN~\cite{jia2021scaling} when there is a one-to-one mapping between an image and its paired caption in a batch, \ie $ \Pcal(i) = \{i\}$ and $ \Qcal(j) = \{ j \} $.

In our empirical study we find that locked pre-trained image and text encoders with trainable gated modules in image encoder work best. Once the customized visual models are trained with the retrieved knowledge, we transfer it to the downstream domain for zero/few/full-shot evaluation.

\vspace{-1mm}
\subsection{Discussions with Data-Centric Methods}
\vspace{-1mm}
It is recommended in~\cite{radford2021learning} that the task learning capabilities of machine learning (ML) systems can be measured by task-level zero-shot transfer. This recommended evaluation setting is further generalized in~\cite{li2022elevater} by showing that few/full-shot transfer consistently yields higher performance than zero-shot transfer. We argue that the task learning capabilities of ML systems can be improved from both the model and data perspectives. Most existing efforts devote to {\it model-centric} methods such as efficient network architectures~\cite{yang2022focal}, smarter training objectives~\cite{yang2022unicl}, and scaling up model size~\cite{goyal2019scaling,yuan2021florence}. {\it Data-centric} methods are less explored, where our retrieval-augmented approach attempts to fill this data gap. We discuss the unique properties of \shortname{} and build the connections with existing data-centric paradigms.

\paragraph{Relation to K-LITE.} To build transferable visual systems, K-LITE~\cite{shen2022klite} enriches entities in language supervision with structural knowledge in  WordNet~\cite{miller1998wordnet} and Wiktionary~\cite{meyer2012wiktionary}, in both model training and evaluation stages. It provides the first strong evidence that structural knowledge is effective in task-level transfer for CLIP/UniCL. Our paper is different in two aspects: 
$(i)$ {\it Knowledge sources.} K-LITE considers textual common sense knowledge bases, while ours considers the web-scale image-text corpus.
$(ii)$ {\it Motivation.} K-LITE aims to improve the generality of visual models via structural human knowledge, while ours improves the customization of visual models using a plug-and-play task instruction augmentation process.

\paragraph{Relation to Self-Training.} As a semi-supervised learning algorithm, self-training~\cite{scudder1965probability,xie2020self} provides pseudo labels to the unlabelled images using a pre-trained neural (teacher) model.
Though sharing the similarity in expanding the task-relevant data, the two methods are different in the augmented knowledge: 
$(i)$ For an image, the supervision signal in self-training is based on the teacher model's internal ``dark knowledge''~\cite{hinton2014dark}, which is limited in a fixed prediction space. The supervision signal in our method is the paired text, which is collected from web as the external knowledge, which may contain richer semantics to describe the image.
$(ii)$ We build a retrieval process to acquire task-relevant images, which is lacking in self-training. The two methods can mutually benefit: self-training can start from our retrieval-augmented pool, while we could use pseudo labels from self-training to get additional supervision.

\vspace{-2mm}
\section{Experiments}
\label{sec:experiments}
\vspace{-2mm}
In this section, we conduct experiments to answer three research questions: (1) What are the unique advantages of retrieval-augmented image-text knowledge for task transfer? (2) How does our design choice of locked-text gated-image tuning compare to existing methods for model customization? (3) Is customization still beneficial in settings where the training data in downstream tasks are observed, \ie, in few-shot or full-shot settings?

We evaluate our models on two CV problems: image classification and image-text retrieval. We first consider ImageNet~\cite{deng2009imagenet} for zero-shot task transfer.  We then further evaluate our model on \textsc{Elevater}~\cite{li2022elevater}, which is an open-set image classification benchmark that contains 20 datasets.  We also conduct experiments on image-text retrieval with MSCOCO~\cite{lin2014microsoft} and Flickr~\cite{young2014image} datasets.

One of the most intriguing benefits of \shortname{} is that it does not need access to any images from the downstream task.  Therefore, we first evaluate on task-level zero-shot transfer, which requires no images in the target to be observed~\cite{radford2021learning,shen2022klite,li2022elevater}. This setting is different from traditional class-level zero-shot~\cite{xian2018zero}, where both the category and images in evaluation should not be observed in training. We argue that ImageNet concepts have been observed in CLIP (Sec.~2.2 of~\cite{radford2021learning}) and other web-scale trained models~\cite{li2021supervision}, as WordNet synsets and common words in English Wikipedia are explicitly added in the query list when searching for (image, text) pairs in their training data construction process.

\vspace{-1mm}
\subsection{Image Classification}
\vspace{-1mm}
\subsubsection{ImageNet-1K}
\vspace{-1mm}

\paragraph{Zero-Shot Task Transfer.}
As shown in Table~\ref{tab:perf_in1k_zeroshot}, by customizing the generic model CLIP/OpenCLIP on 10M retrieved image-text pairs from LAION-400M, \shortname{} achieves a significant and consistent gain (up to 5.4\%) on zero-shot image classification on ImageNet-1K, with different backbones and original pretraining datasets.  There are three interesting findings.

{\it F1: \shortname{} can benefit from model's own pre-training data.}
Compared to OpenCLIP~\cite{openclip} (ViT-B/32) trained on LAION-400M, by training on 10M relevant pairs from the \emph{same} LAION-400M dataset, \shortname{} improves over OpenCLIP by 3.5\%.  Note that the model purely uses the image-text pairs that it has seen during its pre-training, and does \emph{not} see any extra data.  This shows that \shortname{} can more adequately adapt to the target domain during the model customization stage, suggesting a favorable property that no new data is required for customization.

{\it F2: \shortname{} efficiently explores new image-text sources, even for large models.}
We costomize CLIP~\cite{radford2021learning} ViT-L/14 on 10M retrieved relevant image-text pairs, and the model achieves a 2.8\% improvement to 78.1\%.  This surpasses all publicly available checkpoints from CLIP and OpenCLIP, including the checkpoint with a much larger ViT-H/14 backbone and trained on a much larger LAION-2B dataset.  This suggests that \shortname{} is a more sample-efficient approach to improve the model performance on the domain-of-interest.

{\it F3: Scaling up the retrieval pool increases performance.} We perform \shortname{} in a privately collected dataset with over 800M pairs, and train a customized model on 6M retrieved pairs. The performance is increased to 78.5\%, yielding 0.9\% gain compared with 6M pairs retrieved from LAION-400M. This suggests that $\shortname{}$ scales well with the larger retrieval pool.  It showcases  \shortname{} as a cost-efficient approach to leveraging the ever-increasing web image-text corpus.

\begin{table}[t!]
    \centering
    \footnotesize
    \scalebox{0.88}{
    \begin{tabular}{c|l|ll|l|l}
        \toprule
        \multirow{2}{*}{$f_{\thetav}$} & \multirow{2}{*}{Pretrain Data} & \multicolumn{2}{c|}{Retrieved Data } & \multirow{2}{*}{Method} & ImageNet-1K  \\
        & & Dataset  & Size & &  \multicolumn{1}{c}{Zero-Shot}  \\
        \midrule
        \multirow{5}{*}{B/32} & \multirow{2}{*}{WIT-400M} & -- & -- & CLIP & 63.2 \\
        & & L-400M & 10M & \shortname{} & \cellcolor{Gray}{\bf 68.6} {\bf\textcolor{emerald!80}{(+5.4)}}  \\
        \cmidrule{2-6}
        & \multirow{2}{*}{LAION-400M} & -- & -- & OpenCLIP & 62.9 \\  
        & & L-400M & 10M & \shortname{} & \cellcolor{Gray}{\bf 66.4} {\bf\textcolor{emerald!80}{(+3.5)}}\\
        \midrule
        \multirow{6}{*}{L/14} & \multirow{4}{*}{WIT-400M} & -- & -- & CLIP & 75.3 \\
        & & L-400M & 6M & \shortname{} & \cellcolor{Gray}{\bf 77.6} {\bf\textcolor{emerald!80}{(+2.3)}}  \\
        & & L-400M & 10M & \shortname{} & \cellcolor{Gray}{\bf 78.1} {\bf\textcolor{emerald!80}{(+2.8)}}  \\
        & & W-800M$^\dagger$ & 6M & \shortname{} & \cellcolor{Gray}{\bf 78.5} {\bf\textcolor{emerald!80}{(+3.2)}} \\
        \cmidrule{2-6}
        & LAION-400M & -- & -- & OpenCLIP & 72.8 \\
        & LAION-2B & -- & -- & OpenCLIP & 75.3 \\
        \midrule
        H/14 & LAION-2B & -- & -- & OpenCLIP & 78.0 \\
        \bottomrule
    \end{tabular}
    }
    \vspace{-0mm}
    \caption{Comparison of zero-shot task transfer with public checkpoints from CLIP~\cite{radford2021learning} and OpenCLIP~\cite{openclip}.  By continue pretraining on only 10M retrieved data, \shortname{} outperforms \emph{all} public CLIP/OpenCLIP checkpoints, including those with much larger model size and trained on the much larger LAION-2B dataset. LAION~\cite{schuhmann2021laion,schuhmann2022laion} is abbreviated as ``L'' in the table. Web-800M$^\dagger$: a privately collected web database with 800M image-text pairs. Please see the robustness studies on ImageNet dataset variants in Table~\ref{tab:more_perf_in1k_zeroshot}, where \shortname{} consistently outperforms the baseline counterparts in most cases.}
    \label{tab:perf_in1k_zeroshot}
    \vspace{-2mm}
\end{table}

\begin{table}[t]
    \centering
    \footnotesize
    \scalebox{0.88}{
    \begin{tabular}{lll|cc}
        \toprule
        Method & Backbone & \# Params & 1\% & 10\% \\
        \midrule
        \multicolumn{5}{l}{ {\it Self-supervised or semi-supervised methods} } \\
        iBOT~\cite{zhou2021ibot} & ViT-B/16 & 86M & 69.7 & -- \\
        DINO~\cite{assran2021semi} & ViT-B/8 & 86M & 70.0 & -- \\
        MSN~\cite{assran2022masked} & ViT-B/4 & 86M & 75.7 & -- \\
        MSN~\cite{assran2022masked} & ViT-L/7 & 304M & 75.1 & -- \\
        \hline
        PAWS~\cite{assran2021semi} & RN50x4 & 375M & 69.9 & 79.0 \\
        SimCLRv2~\cite{chen2020big} & RN152x3 & 795M & 74.9 & 80.1 \\
        SimCLRv2 (self-distilled)~\cite{chen2020big} & RN152x3 & 795M & 76.6 & 80.9 \\
        Semi-ViT~\cite{cai2022semi} & ViT-H/14 & 632M & 80.0 & 84.3 \\
        SEER~\cite{goyal2021self} & RegNetY &  1.3B & 60.5 & 77.9 \\        
        \midrule
        \multicolumn{5}{l}{ {\it Language-image learning methods} } \\
        CLIP~\cite{radford2021learning} (Zero-Shot) & ViT-B/16 & 86M & \multicolumn{2}{c}{68.6} \\
        CLIP (Random Init.) & ViT-B/16 & 86M & 70.9 & 80.1 \\
        CLIP (Language Init.~\cite{li2022elevater}) & ViT-B/16 & 86M & 74.3 & 80.4 \\
        \rowcolor{Gray}
        \shortname{} (Locked-Text) & ViT-B/16 & 86M & 76.1 & 80.8 \\
        \rowcolor{Gray}
        \shortname{} (Locked-Text Gated-Image) & ViT-B/16 & 129M & 77.4 & 81.8 \\
        \hline
        CLIP~\cite{radford2021learning} (Zero-Shot) & ViT-L/14 & 304M & \multicolumn{2}{c}{75.3} \\
        CLIP (Language Init.\cite{li2022elevater}) & ViT-L/14 & 304M & 80.5 & 84.7 \\
        \rowcolor{Gray}
        \shortname{} (Locked-Text) & ViT-L/14 & 304M & {\bf 81.6} & {\bf 85.1} \\
        \rowcolor{Gray}
        \shortname{} (Locked-Text Gated-Image) & ViT-L/14 & 380M & {\bf 81.6} & {\bf 85.0} \\
        \bottomrule
    \end{tabular}
    }
    \caption{Low-shot (1\% and 10\% labels) on ImageNet-1K. For CLIP and \shortname{} experiments, unless noted, we use language initialization~\cite{li2022elevater} by default for the optimal low-shot performance.
    }
    \label{tab:extreme_lowshot_in1k}
    \vspace{-2mm}
\end{table}

\begin{figure*}[t!]
	\vspace{-0mm}\centering
\includegraphics[height=4.2cm]{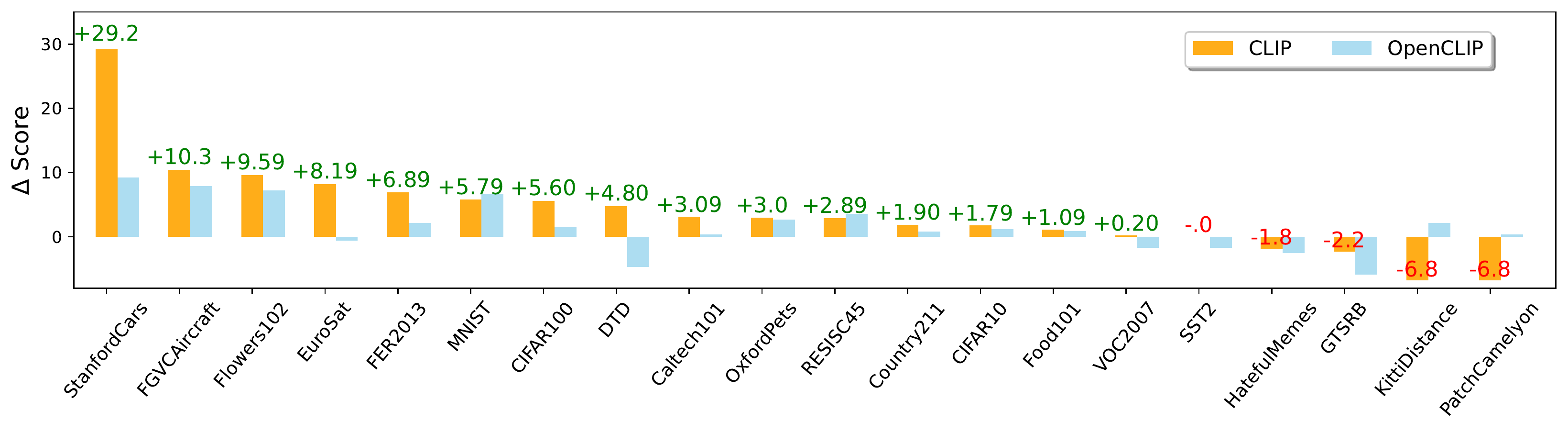} 
	\vspace{-4mm}
	\caption{Zero-shot comparison on \textsc{Elevater} ICinW 20 datasets.  \shortname{} (B32) improves over the base checkpoints on most datasets.
	 }
	\vspace{-3mm}
	\label{fig:icinw_comparison}
\end{figure*}

\paragraph{Low-Shot Adaptation.}
We extend the scope of ImageNet-1K experiments to low-shot settings: 1\% and 10\% labelled data settings, and provide the first strong baselines using CLIP checkpoints.  We present results in Table~\ref{tab:extreme_lowshot_in1k}. First, we find that when the linear head is randomly initialized, it often results in sub-optimal low-shot performance, as the knowledge from the CLIP's language encoder is completely discarded.  We advocate using language-augmented initialization of the linear head~\cite{li2022elevater}, which improves the 1\% label adaptation performance of CLIP ViT-B/16 from 70.9\% to 74.3\% (+3.4\%).  With \shortname{} customization stage, it further improves by 3.1\% to 77.4\%, outperforming prior arts with similar model sizes.  Furthermore, when we further scale up the model size to ViT-L/14, CLIP achieves 80.5\% accuracy, which is on par with the previous SoTA.  \shortname{} further improves the accuracy by 1.1\%, setting a new state-of-the-art of 81.6\% accuracy on 1\% label settings.  Similar trend is observed in 10\% label setting: CLIP is on-par with the prior art, and our \shortname{} customization pushes the new SoTA towards 85.1\% accuracy.

\begin{table}[t!]
    \centering
    \scalebox{0.92}{
    \begin{tabular}{l|c|cc|cc}
    \toprule
        & & \multicolumn{2}{c}{Few-shot} & \multicolumn{2}{c}{Full-shot} \\
        Method & Zero-Shot & LP & FT & LP & FT \\
        \midrule
        CLIP & 56.8 & 65.4 & 63.3 & 78.4 & 80.4 \\
         \rowcolor{Gray}
        \shortname{} & {\bf 60.6} & {\bf 68.9} & {\bf 68.4} & {\bf 80.4} & {\bf 81.8} \\
        \midrule
        Gains
        & {\bf\textcolor{emerald!80}{(+3.8)}} & {\bf\textcolor{emerald!80}{(+3.5)}} & {\bf\textcolor{emerald!80}{(+5.1)}} & {\bf\textcolor{emerald!80}{(+2.0)}} & {\bf\textcolor{emerald!80}{(+1.4)}} \\
        \bottomrule
    \end{tabular}
    }
    \vspace{-0mm}
    \caption{The average scores of image classification performance on 20 datasets in \textsc{Elevater}.  \shortname{} consistently outperforms CLIP in both data-limited and data-rich regimes.}
    \label{tab:perf_elevater_ic}
    \vspace{-3mm}
\end{table}

\begin{figure}[t]
    \begin{subfigure}[h]{\linewidth}
        \centering
    	\includegraphics[width=\linewidth]{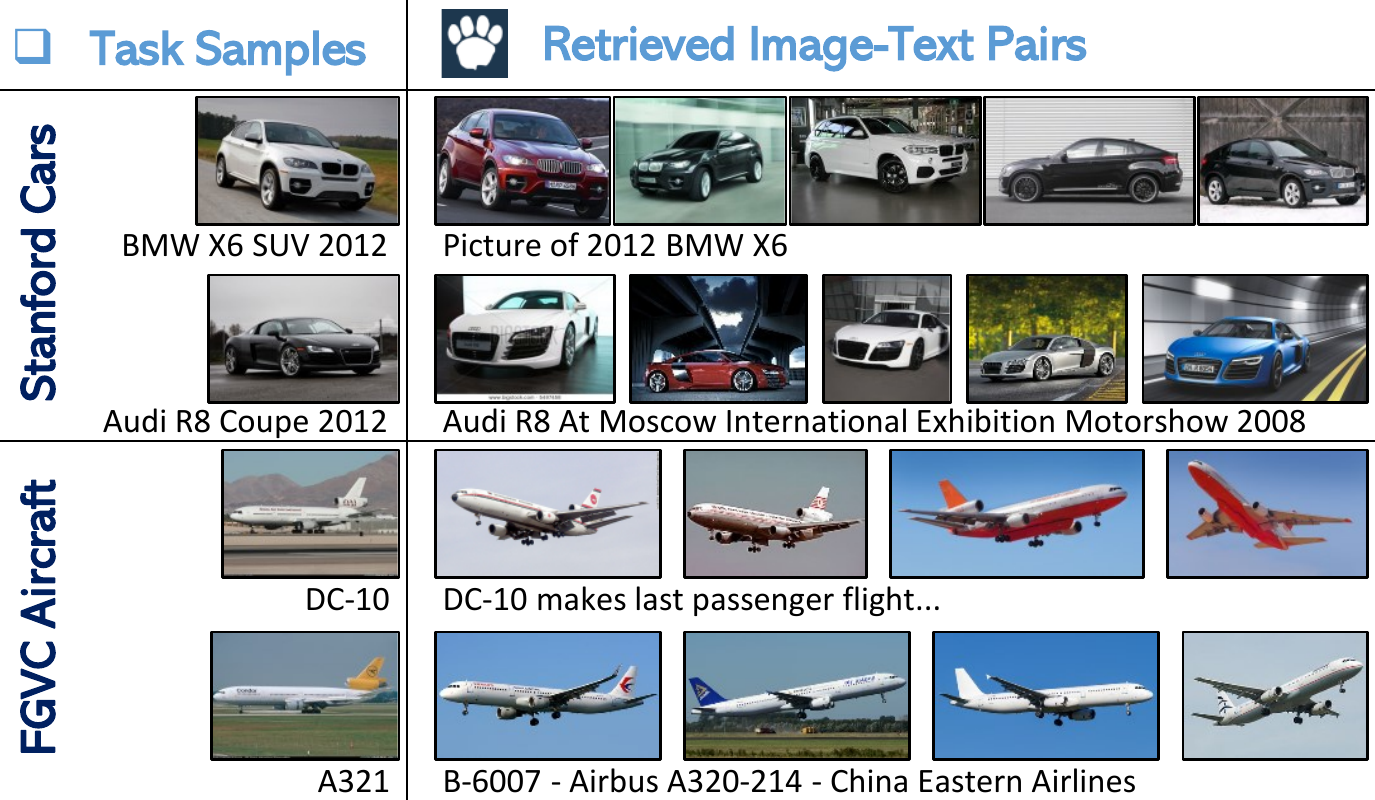}
        \vspace{-2mm}
        \caption{Success examples. The two datasets with largest improvement in Fig.~\ref{fig:icinw_comparison}: Stanford-Cars~\cite{stanfordcars2013} and FGVC-Aircraft~\cite{fgvcaircraft2013}. There is a high concept converage for these datasets in LAION, resulting in a relevant and diverse retrieved set.}
    	\label{fig:success_examples}
    \end{subfigure}
    \begin{subfigure}[h]{\linewidth}
        \centering
    	\includegraphics[width=\linewidth]{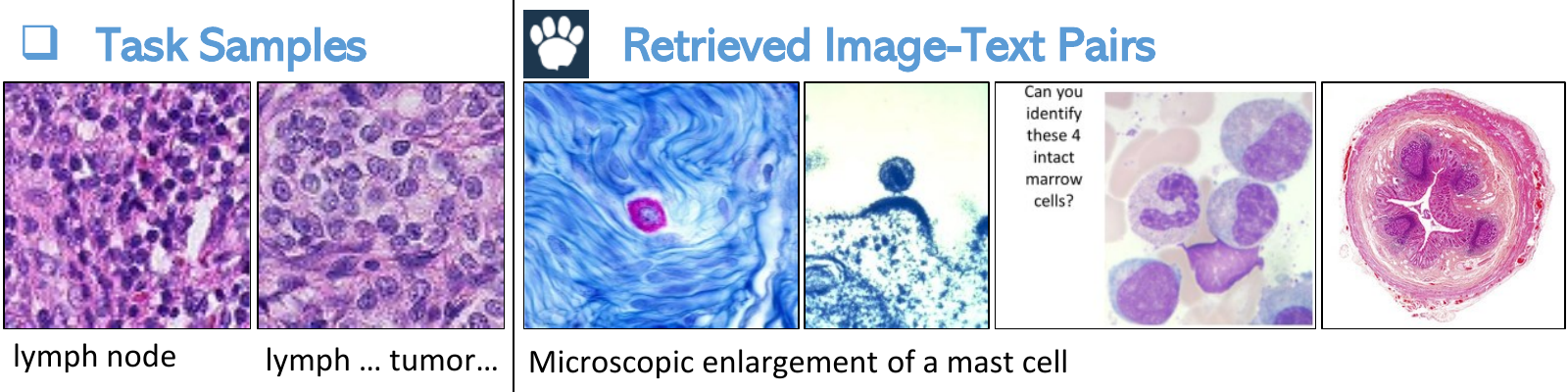}
        \vspace{-2mm}
        \caption{Failure case. The dataset with the largest degradation in Fig.~\ref{fig:icinw_comparison}: PatchCamelyon~\cite{patchcamelyon2018}. LAION-400M has a low concept coverage on this domain, and the retrieved samples are in a  different distribution from the target set.}
    	\label{fig:failure_case}
    \end{subfigure}
    \vspace{-3pt}
    \caption{Success and failure cases in \textsc{Elevater} benchmark.  We show class name and the caption of the first retrieved image-text pairs, others are similar and omitted due to limited space.}
    \label{fig:success_and_failure}
    \vspace{-3mm}
\end{figure}

\vspace{-2mm}
\subsubsection{Zero-, Few-, and Full-Shot on \textsc{Elevater}}
\vspace{-2mm}
As a proxy for performing vision tasks for many customized scenarios in the wild, we consider the {\it image classification in the wild} (ICinW) benchmark in \textsc{Elevater}~\cite{li2022elevater}. It consists of 20 datasets from a diverse selection of domains and covers a wide range of concepts, totaling 1151 classes.

We perform multi-modal knowledge retrieval for 20 datasets together -- the retrieved samples are around 10M image-text pairs in total, on which one single customized visual  model is trained. 
After the process, we feed the customized model to different downstream tasks separately.  For each downstream dataset, we use the official \textsc{Elevater} toolkit to obtain the train/val/test splits, and perform zero-shot, few-shot, and full-shot evaluation.

We report the average scores in Table~\ref{tab:perf_elevater_ic}.
It achieves 3.8\% improvement in the zero-shot setting, even when we do not perform a separate customization for different datasets. This demonstrates the robustness of our customization process.
Further, we see the consistent improvement in few-shot and full-shot settings, including both linear probe (LP) and fine-tuning (FT). This result is encouraging, as it demonstrates that when we have access to some or all data from the downstream task, the proposed model customization stage remains beneficial.  Therefore, \emph{we advocate model customization process in both data-limited and data-rich settings}.

\paragraph{Breakdown Analysis.} Next, we ask why does the retrieved image-text knowledge improve the zero-shot task transfer performance on a broad range of datasets? We compare the breakdown performance on all 20 datasets in Figure~\ref{fig:icinw_comparison} for the zero-shot settings. Out of 20 datasets, the retrieval-augmented knowledge shows superior/comparable/inferior performance to the baseline on 15/1/4 datasets for CLIP and  14/0/6 datasets for OpenCLIP, respectively. Most of the improved and failure datasets are consistent for both checkpoints. For the top two datasets that gains the most, \ie StanfordCars and FGVC Aircraft, relevant image-text knowledge is retrieved from the web-crawled data LAION-400M to describe the concepts; see Fig.~\ref{fig:success_examples}. 
Interestingly, this observation is complementary to K-LITE~\cite{shen2022klite}, which failed on these two datasets, because  no knowledge was extracted from Wiktionary for them, as it often requires domain-specific knowledge and even visual knowledge to best define a car brand (\eg BMW X6 SUV or Audi R8) or an aircraft model type (\eg DC-10 or A321).

\paragraph{Limitations.}
As shown in Fig.~\ref{fig:icinw_comparison}, \shortname{} struggles on the PatchCamelyon dataset, a cancer cell recognition benchmark.  We visualize the retrieved samples and the samples from the original training set in Fig.~\ref{fig:failure_case}. The retrieved images are either instruction photos and from another sensing method, which exhibits a different visual distribution from PatchCamelyon. This suggests the importance of ensuring the retrieval quality for the domain-of-interest.

\begin{table}[t]
    \centering
    \footnotesize
    \scalebox{0.92}{
    \addtolength{\tabcolsep}{-2.8pt}
    \begin{tabular}{@{}p{0.5cm}@{} l| cccc|cccc}
        \toprule
         & \multirow{3}{*}{Method} & \multicolumn{4}{c|}{Flickr30K} &  \multicolumn{4}{c}{MSCOCO} \\
        & & \multicolumn{2}{c}{Img $\rightarrow$ Text} & \multicolumn{2}{c|}{Text $\rightarrow$ Img} & \multicolumn{2}{c}{Img $\rightarrow$ Text} & \multicolumn{2}{c}{Text $\rightarrow$ Img} \\
        & & R@1 & R@5 & R@1 & R@5 & R@1 & R@5 & R@1 & R@5 \\
        \midrule
        \multirow{9}{*}{\rotatebox{90}{Zero-Shot}}
        & ImgBert~\cite{qi2020imagebert} & 70.7 & 90.2 & 54.3 & 79.6 & 44.0 & 71.2 & 32.3 & 59.0 \\
        & ALIGN~\cite{jia2021scaling} & 88.6 & 98.7 & 75.7 & \textbf{93.8} & 58.6 & 83.0 & 45.6 & 69.8 \\
        & CLIP~\cite{radford2021learning} & 88.0 & 98.7 & 68.7 & 90.6 & 58.4 & 81.5 & 37.8 & 62.4 \\
        \cmidrule{2-10}
        & CLIP$^\dagger$ & 87.0 & 98.3 & 66.5 & 88.0 & 59.2 & 80.7 & 37.8 & 62.4 \\
        & \cellcolor{Gray}{ \shortname{}} & \cellcolor{Gray}{\textbf{90.4}} & \cellcolor{Gray}{\textbf{99.1}} & \cellcolor{Gray}{\textbf{76.5}} & \cellcolor{Gray}{\textbf{93.7}} & \cellcolor{Gray}{\textbf{63.3}} & \cellcolor{Gray}{\textbf{85.1}} & \cellcolor{Gray}{\textbf{47.5}} & \cellcolor{Gray}{\textbf{72.0}} \\
        \cmidrule{2-10}
        & Bletchley$^\dagger$ & 90.8 & 98.2 & 78.0 & 94.0 & 66.7 & 85.6 & 48.9 & 72.7 \\
        & \cellcolor{Gray}{ \shortname{}} & \cellcolor{Gray}{\textbf{92.1}} & \cellcolor{Gray}{\textbf{98.7}} & \cellcolor{Gray}{\textbf{79.2}} & \cellcolor{Gray}{\textbf{94.7}} & \cellcolor{Gray}{\textbf{67.7}} & \cellcolor{Gray}{\textbf{85.9}} & \cellcolor{Gray}{\textbf{50.5}} & \cellcolor{Gray}{\textbf{74.4}} \\
        \midrule
        \multirow{4}{*}{\rotatebox{90}{Fine-tuned}} & GPO~\cite{chen2021learning} & 88.7 & 98.9 & 76.1 & 94.5 & 68.1 & 90.2 & 52.7 & 80.2 \\
        & ALIGN~\cite{jia2021scaling} & 95.3 & \textbf{99.8} & 84.9 & 97.4 & 77.0 & 93.5 & 59.9 & 83.3 \\
        & CLIP$^\dagger$ & 96.4 & \textbf{99.8} & 86.5 & \textbf{97.9} & 78.3 & 93.8 & 60.9 & 83.8 \\
       
        & \cellcolor{Gray}{\shortname{}} & \cellcolor{Gray}{\textbf{96.6}} & \cellcolor{Gray}{\textbf{99.9}} & \cellcolor{Gray}{\textbf{86.8}} & \cellcolor{Gray}{\textbf{98.0}} & \cellcolor{Gray}{\textbf{78.7}} & \cellcolor{Gray}{\textbf{94.0}} & \cellcolor{Gray}{\textbf{61.1}} & \cellcolor{Gray}{\textbf{84.1}} \\
        \bottomrule
    \end{tabular}
    \addtolength{\tabcolsep}{2.8pt}
    }
    \vspace{-0mm}
    \caption{Image-text retrieval results on Flickr30K~\cite{plummer2015flickr30k} and MSCOCO~\cite{lin2014microsoft} datasets. CLIP$^\dagger$, Bletchley$^\dagger$: our evaluation.}
    \label{tab:perf_retrieval}
    \vspace{-3mm}
\end{table}

\subsection{Image-Text Retrieval}

\label{sec:image-text-retrieval}

To demonstrate the generality of \shortname{}, we consider Flickr30K~\cite{young2014image} and MSCOCO~\cite{lin2014microsoft} image-text retrieval tasks, in both zero-shot and full-shot settings. 
We use the standard image-text contrastive objective~\cite{radford2021learning}.  For image-text retrieval task, following \cite{radford2021learning,jia2021scaling}, we use the CLIP-L/14 with 336x336 input resolution in both zero-shot, customization, and fine-tuning stage.
We use the captions from MSCOCO as queries to retrieve 6M image-text pairs and perform customization.  Note that \emph{none} of the caption queries are used in the model training stage.

As shown in Table~\ref{tab:perf_retrieval}, \shortname{} improves the generic CLIP counterparts on both zero-shot and full-shot retrieval for Flickr30K and MSCOCO datasets.
The gain on zero-shot task transfer is large. 
On Flickr30K, it achieves 3.4\%/10.0\% recall improvement for I2T and T2I retrieval, respectively. Afer fine-tuning on full training data, \shortname{} still improves over the baseline slightly.
It provides another piece of evidence for \shortname{} in data-rich settings.
Furthermore, we conduct the same customization procedure of \shortname{} on a large checkpoint Bletchley~\cite{bletchley} with 864M parameters, and observe consistent gains over both datasets. It demonstrates that \shortname{} scales well with model size on retrieval tasks.

\subsection{Dense Prediction Tasks}
\label{sec:dense_prediction}

Although \shortname{} is optimized with the image-level contrastive loss during the customization stage, we find it beneficial for dense prediction tasks as well.  We showcase its application to dense prediction tasks on object detection and semantic segmentation.

\subsubsection{Object Detection}
\label{sec:object_detection}

For object detection, we choose the state-of-the-art RegionCLIP~\cite{zhong2021regionclip} as our framework.  We conduct experiments in two settings: zero-shot inference with ground-truth (GT)/ Region Proposal Networks (RPN) boxes and open-vocabulary object detection on MSCOCO dataset.  We perform the model customization following the same setting as Sec.~\ref{sec:image-text-retrieval}.  Following RegionCLIP, we conduct experiments on ResNet50 backbone.  Additionally, we present results on ViT-B/16 backbone for zero-shot inference.

\paragraph{Zero-shot inference.}
RegionCLIP alters the architecture of an object detector. It is able to perform zero-shot object detection, by (1) initializing its backbone and prediction head from the CLIP-pretrained checkpoints, and  (2) employing a pretrained RPN network.
The results are shown in Table~\ref{tab:perf_od}. \shortname{} consistently improves over CLIP checkpoint under all settings.
When ground-truth region proposal is used, \shortname{} improves over CLIP by +1.0 on overall AP50; when the pretrained RPN is used, \shortname{} demonstrates +1.5/+1.4/+1.9 AP50 improvements on novel, base, and all classes, respectively.
These results are encouraging, as it shows that the customized knowledge from \shortname{} transfers well to dense prediction tasks like object detection, under the RegionCLIP framework.

\begin{table}[t]
    \centering
    \footnotesize
    \scalebox{0.9}{
    \addtolength{\tabcolsep}{-2.5pt}
    \begin{tabular}{lll|c|lll}
        \toprule
         & Pretrain & \multirow{2}{*}{Backbone} & Region & \multicolumn{3}{c}{MSCOCO AP$_{50}$} \\
        & Method & & Proposals & Novel & Base & All \\
        \midrule
        \multirow{7}{*}{\rotatebox{90}{Zero-Shot}} & CLIP & ResNet-50 & GT & 58.6 & 58.2 & 58.3 \\
        & \cellcolor{Gray}{\shortname{}} & \cellcolor{Gray}{ResNet-50} & \cellcolor{Gray}{GT} & \cellcolor{Gray}{\textbf{58.9} {\bf\textcolor{emerald!80}{(+0.3)}}} & \cellcolor{Gray}{\textbf{59.4} {\bf\textcolor{emerald!80}{(+1.2)}}} & \cellcolor{Gray}{\textbf{59.3} {\bf\textcolor{emerald!80}{(+1.0)}}} \\
        \cmidrule{2-7}
        & CLIP & ResNet-50 & RPN & 29.7 & 24.0 & 25.5 \\
        & \cellcolor{Gray}{\shortname{}} & \cellcolor{Gray}{ResNet-50} & \cellcolor{Gray}{RPN} & \cellcolor{Gray}{\textbf{31.6} {\bf\textcolor{emerald!80}{(+1.9)}}} & \cellcolor{Gray}{\textbf{25.4} {\bf\textcolor{emerald!80}{(+1.4)}}} & \cellcolor{Gray}{\textbf{27.0} {\bf\textcolor{emerald!80}{(+1.5)}}} \\
        \cmidrule{2-7}
        & CLIP & ViT-B/16 & GT & 40.2 & 38.5 & 38.9 \\
        & \cellcolor{Gray}{\shortname{}} & \cellcolor{Gray}{ViT-B/16} & \cellcolor{Gray}{GT} & \cellcolor{Gray}{\textbf{43.5} {\bf\textcolor{emerald!80}{(+3.3)}}} & \cellcolor{Gray}{\textbf{47.2} {\bf\textcolor{emerald!80}{(+8.7)}}} & \cellcolor{Gray}{\textbf{46.3} {\bf\textcolor{emerald!80}{(+7.4)}}} \\
        \midrule
        \multirow{2}{*}{\rotatebox{90}{OVD}} & CLIP & ResNet-50 & -- & 14.2 & 52.8 & 42.7 \\
        & \cellcolor{Gray}{\shortname{}} & \cellcolor{Gray}{ResNet-50} & \cellcolor{Gray}{--} & \cellcolor{Gray}{\textbf{20.6} {\bf\textcolor{emerald!80}{(+6.4)}}} & \cellcolor{Gray}{\textbf{55.1} {\bf\textcolor{emerald!80}{(+2.3)}}} & \cellcolor{Gray}{\textbf{46.1} {\bf\textcolor{emerald!80}{(+3.4)}}} \\
        \bottomrule
    \end{tabular}
    \addtolength{\tabcolsep}{2.5pt}
    }
    \caption{Zero-shot and open-vocabulary object detection results on MSCOCO~\cite{lin2014microsoft} dataset using RegionCLIP~\cite{zhong2021regionclip} pipeline.}
    \label{tab:perf_od}
\end{table}

\paragraph{Open-vocabulary detection (OVD).}
We further conduct experiments on the open-vocabulary settings, where the model finetunes on a set of selected categories (\emph{base} classes), and evaluate on both seen (\emph{base}) and unseen (\emph{novel}) classes.  We report the results in Table~\ref{tab:perf_od} (OVD).

We can see that with the \shortname{} customization, the detector yields improved performance on base with +2.3 AP50, and importantly, it significantly improves novel categories with +6.4 AP50.  This suggests that the injected knowledge during the model customization stage improves the learned fine-grained visual feature that is beneficial to both seen and unseen categories for object detection, when the downstream coarse-grained data is available. This is favored, because (1) the weakly-supervised data such as the coarse-grained image-text pairs requires much less human annotation cost than fine-grained bounding box annotation, (2) the paired data in \shortname{} is free, as it is retrieved from the web, where COCO image-text pairs are not used in customized training.

\begin{table}[t]
    \centering
    \footnotesize
    \scalebox{1.0}{
    \begin{tabular}{l|l}
        \toprule
         Method & mIoU \\
        \midrule
        MaskCLIP~\cite{zhou2022maskclip} & 12.5 \\
        \cellcolor{Gray}{\shortname{} (Locked-Text)} & \cellcolor{Gray}{\bf 14.4 \textcolor{emerald!80}{(+1.9)}} \\
        \cellcolor{Gray}{\shortname{} (Locked-Text Gated-Image)} & \cellcolor{Gray}{\bf 14.5 \textcolor{emerald!80}{(+2.0)}} \\
        \midrule
        \multicolumn{2}{l}{{\it w/ key smoothing and prompt denoising}} \\
        MaskCLIP~\cite{zhou2022maskclip} & 14.6 \\
        \cellcolor{Gray}{\shortname{} (Locked-Text)} & \cellcolor{Gray}{\bf 18.2 \textcolor{emerald!80}{(+3.6)}} \\
        \cellcolor{Gray}{\shortname{} (Locked-Text Gated-Image)} & \cellcolor{Gray}{\bf 16.3 \textcolor{emerald!80}{(+1.7)}} \\
        \midrule
        \multicolumn{2}{l}{{\it w/ full-shot finetuning using pseudo labels}} \\
        MaskCLIP+~\cite{zhou2022maskclip} & 18.1 \\
        \cellcolor{Gray}{\shortname{} (Locked-Text)} & \cellcolor{Gray}{\bf 20.7 \textcolor{emerald!80}{(+2.6)}} \\
        \cellcolor{Gray}{\shortname{} (Locked-Text Gated-Image)} & \cellcolor{Gray}{\bf 19.2 \textcolor{emerald!80}{(+1.1)}} \\
        \bottomrule
    \end{tabular}
    }
    \caption{Zero-shot and annotation-free semantic segmentation results on COCO Stuff~\cite{lin2014microsoft} using MaskCLIP~\cite{zhou2022maskclip} (ViT-B/16). }
    \label{tab:perf_zeroshot_seg}
\end{table}

\subsubsection{Semantic Segmentation}

\label{sec:semantic_segmentation}

For semantic segmentation, we choose the state-of-the-art MaskCLIP~\cite{zhou2022maskclip} as the framework. It investigates three evaluation settings for segmentation. First, it makes use of the pretrained CLIP checkpoint to discover the alignment between grid visual features and the text prompt features, so as to perform zero-shot semantic segmentation.
Second, to further improve the performance, MaskCLIP~\cite{zhou2022maskclip} proposes two techniques for refining its zero-shot predictions: key smoothing and prompt denoising. 
Lastly, when the training images are available, without the need to access the training labels, it further proposes MaskCLIP+~\cite{zhou2022maskclip} to perform full-shot finetuning on the target training set using pseudo-labels.
Following MaskCLIP~\cite{zhou2022maskclip}, we use ViT-B/16 checkpoints, and use their official code base to train and evaluate models.  We report results in Table~\ref{tab:perf_zeroshot_seg}.

On all of the three settings, \shortname{} demonstrates improvements over the MaskCLIP, with both locked-text and lock-text-gated-image tuning.  This indicates that \shortname{}-based model customization is beneficial for dense prediction tasks as well.  Notably, when key smoothing and prompt denoising is used, \shortname{} with locked-text tuning demonstrates a significant 3.6\% gain in mIoU.
Besides, the locked-text tuning benefit more than gated-image tuning, when prediction refinement techniques are used.  We hypothesize that locked-text tuning may generate more noisy predictions comparing with gated-image predictions, and it can thus benefit more with the proposed refinement techniques.

Surprisingly, without seeing the downstream COCO images, the zero-shot evaluation of \shortname{} (18.2) even slightly outperforms MaskCLIP+ (18.1), which is finetuned on the downstream training COCO images with self-training.

\subsection{Ablation Studies}

\label{sec:ablation}

We ablate \shortname{} on ImageNet with CLIP ViT-B/32 checkpoint, with 10M retrieved image-text pairs from LAION-400M.

\begin{figure}[t]
	\centering
	\includegraphics[width=0.85\linewidth]{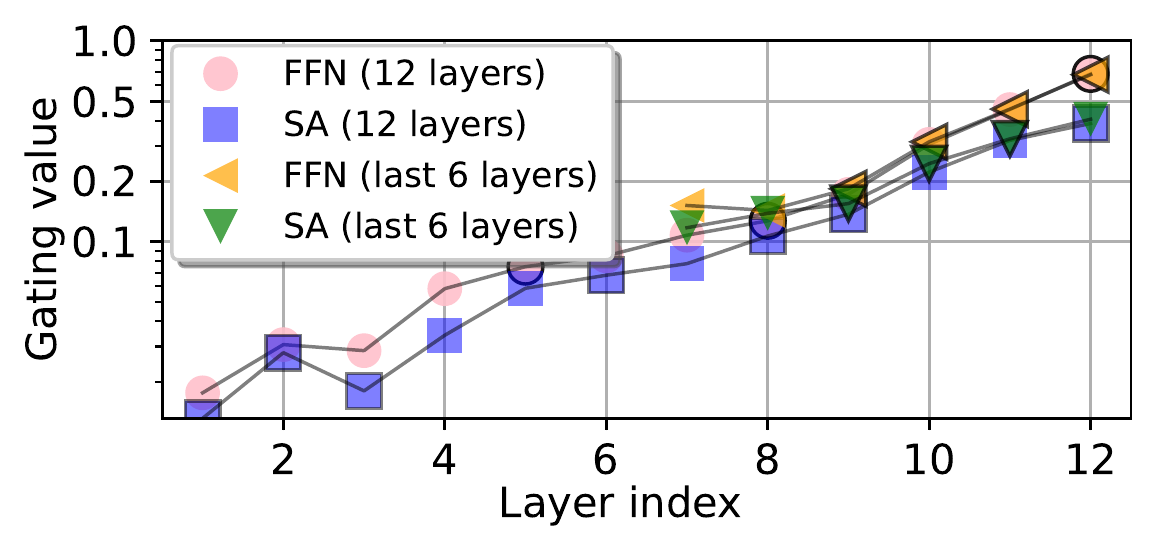}\\
	\vspace{-3mm}
	\includegraphics[width=0.85\linewidth]{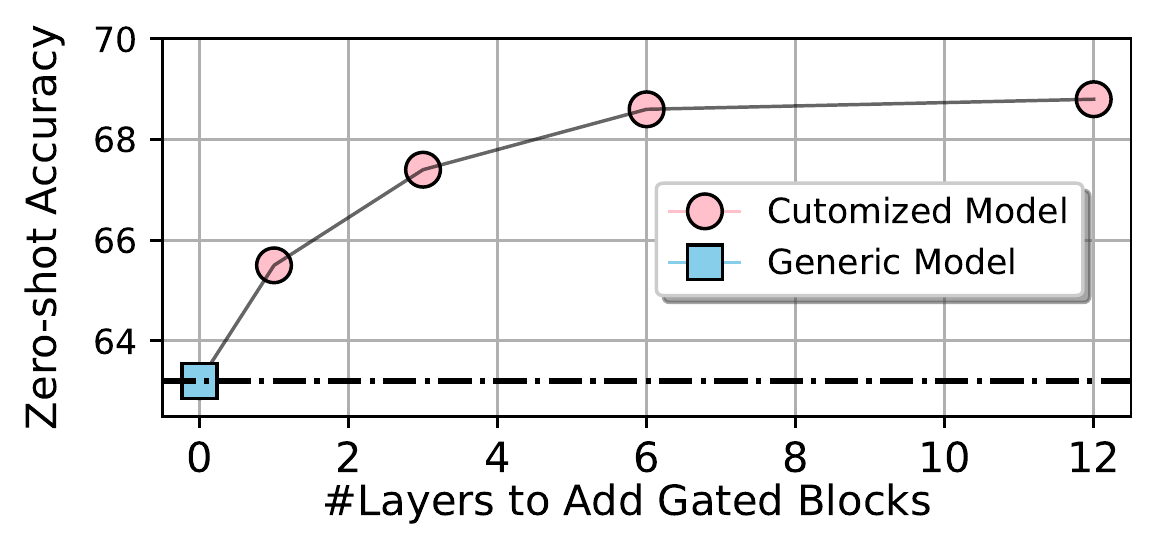}
    \vspace{-3mm}
    \caption{(Top) The learned gated values on the newly added SA and FFN layers for two networks: adding gated blocks in all 12 layers and in the last 6 layers, respectively; (Bottom) the zero-shot accuracy on ImageNet when adding gated blocks into different number of last layers.}
    \label{fig:gated_blocks}
    \vspace{-3mm}
\end{figure}

\paragraph{Where to add gated blocks?}  We conduct an experiment by adding a single GSA block before each Transformer block and continual pretraining the model on the retrieved image-text pairs.  We then visualize the learned gated values in Fig.~\ref{fig:gated_blocks} (top): as the network goes deeper, the gate values become larger, and compared with other blocks, the learned alpha gates in the first six layers have a much smaller value.  We hypothesize that earlier blocks have a smaller modulation to the base network, and removing them has minimal effect on the model's performance. Therefore, we vary the number of \emph{last} layers which we add gated blocks to, and show empirical results in Fig.~\ref{fig:gated_blocks} (bottom).  With more layers added to the base network, the model's zero-shot performance gradually increases, and saturates at 6 blocks.
Therefore, we add gated blocks to last 6 layers as the default to balance the accuracy and the efficiency.

\paragraph{Tuning strategy.}  We ablate the design choices in the model customization stage: (1) direct tuning the pre-trained weights \vs training gated blocks from scratch; (2) updating visual \vs text encoder.  We report results in  Table~\ref{tab:ablation_finetune_strategy}.
First, a frozen text encoder consistently outperforms a frozen visual encoder. We conjecture the phenomenon is due to that the retrieved texts have a much more limited space, comparing to text space in the original pre-training set (\eg LAION-400M), as the concepts are limited to the query classes from the target domain. 
Therefore, fine-tuning the text encoder may tend to collapse the pre-trained semantic space.

\begin{table}[t]
    \centering
    \footnotesize
    \scalebox{0.83}{
    \begin{tabular}{@{}p{0.3cm}c p{0.4cm} p{0.4cm} p{0.9cm} c cc}
        & & & & & IN1K & \multicolumn{2}{c}{COCO R@1} \\
        & Method & Visual & Text & \# Train. & Acc. & T2I & I2T \\
        \shline
        \multirow{4}{*}{\rotatebox{90}{Direct}}  
        & CLIP~\cite{radford2021learning}  & \xmark & \xmark & -- & 63.2 & 48.8 & 29.9 \\
        & Locked-Image~\cite{zhai2022lit} & \xmark & \cmark & 63.4M & 63.7 & 50.5 & 34.2 \\
        & \cellcolor{Gray}{Locked-Text} & \cellcolor{Gray}{\cmark} & \cellcolor{Gray}{\xmark} & \cellcolor{Gray}{88.1M} & \cellcolor{Gray}{66.9} & \cellcolor{Gray}{51.1} & \cellcolor{Gray}{36.1} \\
        & Full-model & \cmark & \cmark & 151.3M & 64.3 & 54.3 & 37.9 \\
        \hline
        \multirow{3}{*}{\rotatebox{90}{Gated}}  
        & & \xmark & \cmark & 18.9M & 63.0 & 49.5 & 33.5 \\
        & \cellcolor{Gray}{Locked-Text Gated-Image} & \cellcolor{Gray}{\cmark} & \cellcolor{Gray}{\xmark} & \cellcolor{Gray}{42.5M} & \cellcolor{Gray}{\textbf{68.6}} & \cellcolor{Gray}{53.4} & \cellcolor{Gray}{38.1} \\
        &  & \cmark & \cmark & 89.8M & \textbf{68.7} & 54.3 & 39.9 \\
    \end{tabular}
    }
    \vspace{-2mm}
    \caption{Ablation: tuning strategy. For our model customization purpose, we advocate locked-text (gated-image) tuning methods in gray rows. \cmark: trainable, \xmark: locked.}
    \label{tab:ablation_finetune_strategy}
    \vspace{-3mm}
\end{table}

We advocate two tuning methods for model customization. 
Locked-text gated-image tuning has a strong adaptation power, and is efficient in the model customization stage, with fewer trainable parameters. Locked-text tuning is also an effective way of customizing the models to downstream tasks, without the need of adding extra parameters.  By default, we use gated blocks for its superior performance and efficiency.

\paragraph{Retrieval size.}   We observe that training with a small retrieval size is more likely to overfit. We find that a larger retrieval size generally yields better performance, and saturates at around 6-10M image-text pairs. 
\begin{table}[h!]
    \centering
    \footnotesize
    \begin{tabular}{r|ccccc}
        Retrieval Size & 0 & 1M & 3M & 6M & 10M \\
        \shline
        ImageNet-1K Accuracy & 63.2 & 64.8 & 66.9 & \textbf{68.6} & \textbf{68.6} \\
    \end{tabular}
    \vspace{-3mm}
\end{table}

\paragraph{Search methods.}  Three search methods are compared: T2I, T2T, and T2I/T2T-combined.  We find all modes consistently improve over the baseline CLIP.  T2I retrieval alone yields a slightly worse performance, which may be partly due to its retrieved samples being more noisy and less relevant than T2T retrieval.  T2T alone or T2I/T2T-combined has a similar performance.  We use T2I/T2T-combined as our default strategy.

\begin{table}[h]
    \centering
    \footnotesize
    \begin{tabular}{r|cccc}
        Search Methods & -- & T2I & T2T & T2I/T2T \\
        \shline
        ImageNet-1K Accuracy & 63.2 & 65.8 & \textbf{68.6} & \textbf{68.6} \\
    \end{tabular}
    \vspace{-3mm}
\end{table}

\paragraph{Comparison with self-training.}  We study and compare with the pseudo labeling strategy on retrieval-augmented model customization.  After the relevant image-text pairs are retrieved, we use the pretrained CLIP checkpoint to assign pseudo labels to each retrieved image, and finetune the model as a classification task using the retrieved samples and pseudo labels.  We optimize the network with UniCL loss~\cite{yang2022unicl}.  With frozen text, pseudo-labeling can improve the model's performance with T2I/T2T data, while having a decreased performance with T2I data. This can be due to that the relevance of the T2I data is lower than other splits, and pseudo label assigned can be incorrect.  Besides, gated self-attention is complimentary to pseudo labeling with $\sim$2\% improvements.
In contrast to pseudo labeling, training directly on the retrieved image-text pairs does not use heuristics to create pseudo labels, and the model can receive additional supervision signal, which we empirically find helpful for model adaptation and robustness.

\begin{table}[h!]
\vspace{-2mm}
    \centering
    \footnotesize
    \begin{tabular}{l|l|ccc}
        & & \multicolumn{3}{c}{Retrieval Methods} \\
        Method & Tuning Strategy & T2I & T2T & T2I/T \\
        \shline
        CLIP & -- & & 63.2 & \\
        \hline
        Self-training & Locked-Text & 62.4 & 63.6 & 64.6 \\
        & Locked-Text Gated-Image & 64.3 & 66.1 & 66.2 \\
        \hline
        \shortname{} & Locked-Text Gated-Image & \textbf{65.8} & \textbf{68.6} & \textbf{68.6} \\
    \end{tabular}
    \vspace{-0mm}
    \label{tab:ablation_pseudo_label}
    \vspace{-3mm}
\end{table}

\section{Conclusion}
\label{sec:conclusion}

We presented \shortname{}, a plug-and-play framework for leveraging large-scale image-text corpus as external knowledge to efficiently customize models on downstream tasks.  Extensive experiments demonstrate its generality and effectiveness in image classification, image-text retrieval, object detection, and semantic segmentation, on more than 20 different downstream datasets.  We highly advocate the model customization stage for building more transferable visual system for different downstream tasks.

\paragraph{Acknowledgement.}
We thank Min Gao, Ping Jin, Houdong Hu from Microsoft Azure AI for LAION data preprocessing and evaluation scripts, Qiang Liu and Subhojit Som from Microsoft Turing Team for the Bletchley model checkpoints.
This work was supported in part by NSF CAREER IIS2150012, NASA 80NSSC21K0295, and Institute of Information \& communications Technology Planning \& Evaluation(IITP) grants funded by the Korea government(MSIT) (No. 2022- 0-00871, Development of AI Autonomy and Knowledge Enhancement for AI Agent Collaboration) and (No. RS-2022-00187238, Development of Large Korean Language Model Technology for Efficient Pre-training).

{\small
\bibliographystyle{ieee_fullname}
\bibliography{egbib}

\begin{thebibliography}{10}\itemsep=-1pt

\bibitem{bletchley}
Turing {B}letchley.
\newblock
  \url{https://www.microsoft.com/en-us/research/blog/turing-bletchley-a-universal-image-language-representation-model-by-microsoft/}.

\bibitem{alayrac2022flamingo}
Jean-Baptiste Alayrac, Jeff Donahue, Pauline Luc, Antoine Miech, Iain Barr,
  Yana Hasson, Karel Lenc, Arthur Mensch, Katie Millican, Malcolm Reynolds,
  et~al.
\newblock Flamingo: a visual language model for few-shot learning.
\newblock {\em arXiv preprint arXiv:2204.14198}, 2022.

\bibitem{assran2022masked}
Mahmoud Assran, Mathilde Caron, Ishan Misra, Piotr Bojanowski, Florian Bordes,
  Pascal Vincent, Armand Joulin, Mike Rabbat, and Nicolas Ballas.
\newblock Masked siamese networks for label-efficient learning.
\newblock In {\em European Conference on Computer Vision}, pages 456--473.
  Springer, 2022.

\bibitem{assran2021semi}
Mahmoud Assran, Mathilde Caron, Ishan Misra, Piotr Bojanowski, Armand Joulin,
  Nicolas Ballas, and Michael Rabbat.
\newblock Semi-supervised learning of visual features by non-parametrically
  predicting view assignments with support samples.
\newblock In {\em Proceedings of the IEEE/CVF International Conference on
  Computer Vision}, pages 8443--8452, 2021.

\bibitem{blattmann2022retrieval}
Andreas Blattmann, Robin Rombach, Kaan Oktay, and Bj{\"o}rn Ommer.
\newblock Retrieval-augmented diffusion models.
\newblock {\em arXiv preprint arXiv:2204.11824}, 2022.

\bibitem{borgeaud2021improving}
Sebastian Borgeaud, Arthur Mensch, Jordan Hoffmann, Trevor Cai, Eliza
  Rutherford, Katie Millican, George van~den Driessche, Jean-Baptiste Lespiau,
  Bogdan Damoc, Aidan Clark, et~al.
\newblock Improving language models by retrieving from trillions of tokens.
\newblock {\em arXiv preprint arXiv:2112.04426}, 2021.

\bibitem{cai2022semi}
Zhaowei Cai, Avinash Ravichandran, Paolo Favaro, Manchen Wang, Davide Modolo,
  Rahul Bhotika, Zhuowen Tu, and Stefano Soatto.
\newblock Semi-supervised vision transformers at scale.
\newblock {\em arXiv preprint arXiv:2208.05688}, 2022.

\bibitem{caron2021emerging}
Mathilde Caron, Hugo Touvron, Ishan Misra, Herv{\'e} J{\'e}gou, Julien Mairal,
  Piotr Bojanowski, and Armand Joulin.
\newblock Emerging properties in self-supervised vision transformers.
\newblock {\em ICCV}, 2021.

\bibitem{changpinyo2021conceptual}
Soravit Changpinyo, Piyush Sharma, Nan Ding, and Radu Soricut.
\newblock Conceptual 12m: Pushing web-scale image-text pre-training to
  recognize long-tail visual concepts.
\newblock In {\em CVPR}, 2021.

\bibitem{chen2021learning}
Jiacheng Chen, Hexiang Hu, Hao Wu, Yuning Jiang, and Changhu Wang.
\newblock Learning the best pooling strategy for visual semantic embedding.
\newblock In {\em Proceedings of the IEEE/CVF conference on computer vision and
  pattern recognition}, pages 15789--15798, 2021.

\bibitem{chen2020simple}
Ting Chen, Simon Kornblith, Mohammad Norouzi, and Geoffrey Hinton.
\newblock A simple framework for contrastive learning of visual
  representations.
\newblock In {\em ICML}, 2020.

\bibitem{chen2020big}
Ting Chen, Simon Kornblith, Kevin Swersky, Mohammad Norouzi, and Geoffrey~E
  Hinton.
\newblock Big self-supervised models are strong semi-supervised learners.
\newblock {\em Advances in neural information processing systems},
  33:22243--22255, 2020.

\bibitem{chen2022murag}
Wenhu Chen, Hexiang Hu, Xi Chen, Pat Verga, and William~W Cohen.
\newblock Murag: Multimodal retrieval-augmented generator for open question
  answering over images and text.
\newblock {\em arXiv preprint arXiv:2210.02928}, 2022.

\bibitem{chen2022re}
Wenhu Chen, Hexiang Hu, Chitwan Saharia, and William~W Cohen.
\newblock Re-imagen: Retrieval-augmented text-to-image generator.
\newblock {\em arXiv preprint arXiv:2209.14491}, 2022.

\bibitem{chen2022pali}
Xi Chen, Xiao Wang, Soravit Changpinyo, AJ Piergiovanni, Piotr Padlewski,
  Daniel Salz, Sebastian Goodman, Adam Grycner, Basil Mustafa, Lucas Beyer,
  et~al.
\newblock Pali: A jointly-scaled multilingual language-image model.
\newblock {\em arXiv preprint arXiv:2209.06794}, 2022.

\bibitem{deng2009imagenet}
Jia Deng, Wei Dong, Richard Socher, Li-Jia Li, Kai Li, and Li Fei-Fei.
\newblock Imagenet: A large-scale hierarchical image database.
\newblock In {\em CVPR}, 2009.

\bibitem{dong2022maskclip}
Xiaoyi Dong, Yinglin Zheng, Jianmin Bao, Ting Zhang, Dongdong Chen, Hao Yang,
  Ming Zeng, Weiming Zhang, Lu Yuan, Dong Chen, et~al.
\newblock Maskclip: Masked self-distillation advances contrastive
  language-image pretraining.
\newblock {\em arXiv preprint arXiv:2208.12262}, 2022.

\bibitem{dosovitskiy2020image}
Alexey Dosovitskiy, Lucas Beyer, Alexander Kolesnikov, Dirk Weissenborn,
  Xiaohua Zhai, Thomas Unterthiner, Mostafa Dehghani, Matthias Minderer, Georg
  Heigold, Sylvain Gelly, et~al.
\newblock An image is worth 16x16 words: Transformers for image recognition at
  scale.
\newblock {\em arXiv preprint arXiv:2010.11929}, 2020.

\bibitem{gao2022pyramidclip}
Yuting Gao, Jinfeng Liu, Zihan Xu, Jun Zhang, Ke Li, and Chunhua Shen.
\newblock Pyramidclip: Hierarchical feature alignment for vision-language model
  pretraining.
\newblock {\em arXiv preprint arXiv:2204.14095}, 2022.

\bibitem{geng2022multimodal}
Xinyang Geng, Hao Liu, Lisa Lee, Dale Schuurams, Sergey Levine, and Pieter
  Abbeel.
\newblock Multimodal masked autoencoders learn transferable representations.
\newblock {\em arXiv preprint arXiv:2205.14204}, 2022.

\bibitem{goyal2021self}
Priya Goyal, Mathilde Caron, Benjamin Lefaudeux, Min Xu, Pengchao Wang, Vivek
  Pai, Mannat Singh, Vitaliy Liptchinsky, Ishan Misra, Armand Joulin, et~al.
\newblock Self-supervised pretraining of visual features in the wild.
\newblock {\em arXiv preprint arXiv:2103.01988}, 2021.

\bibitem{goyal2019scaling}
Priya Goyal, Dhruv Mahajan, Abhinav Gupta, and Ishan Misra.
\newblock Scaling and benchmarking self-supervised visual representation
  learning.
\newblock In {\em ICCV}, 2019.

\bibitem{guu2020realm}
Kelvin Guu, Kenton Lee, Zora Tung, Panupong Pasupat, and Ming-Wei Chang.
\newblock Realm: Retrieval-augmented language model pre-training.
\newblock {\em arXiv preprint arXiv:2002.08909}, 2020.

\bibitem{he2022masked}
Kaiming He, Xinlei Chen, Saining Xie, Yanghao Li, Piotr Doll{\'a}r, and Ross
  Girshick.
\newblock Masked autoencoders are scalable vision learners.
\newblock In {\em Proceedings of the IEEE/CVF Conference on Computer Vision and
  Pattern Recognition}, pages 16000--16009, 2022.

\bibitem{he2020momentum}
Kaiming He, Haoqi Fan, Yuxin Wu, Saining Xie, and Ross Girshick.
\newblock Momentum contrast for unsupervised visual representation learning.
\newblock In {\em CVPR}, 2020.

\bibitem{hendrycks2021many}
Dan Hendrycks, Steven Basart, Norman Mu, Saurav Kadavath, Frank Wang, Evan
  Dorundo, Rahul Desai, Tyler Zhu, Samyak Parajuli, Mike Guo, Dawn Song, Jacob
  Steinhardt, and Justin Gilmer.
\newblock The many faces of robustness: A critical analysis of
  out-of-distribution generalization.
\newblock {\em ICCV}, 2021.

\bibitem{hendrycks2021nae}
Dan Hendrycks, Kevin Zhao, Steven Basart, Jacob Steinhardt, and Dawn Song.
\newblock Natural adversarial examples.
\newblock {\em CVPR}, 2021.

\bibitem{hinton2014dark}
Geoffrey Hinton, Oriol Vinyals, and Jeff Dean.
\newblock Dark knowledge.
\newblock {\em Presented as the keynote in BayLearn}, 2014.

\bibitem{openclip}
Gabriel Ilharco, Mitchell Wortsman, Ross Wightman, Cade Gordon, Nicholas
  Carlini, Rohan Taori, Achal Dave, Vaishaal Shankar, Hongseok Namkoong, John
  Miller, Hannaneh Hajishirzi, Ali Farhadi, and Ludwig Schmidt.
\newblock Openclip.
\newblock July 2021.
\newblock If you use this software, please cite it as below.

\bibitem{jain2021mural}
Aashi Jain, Mandy Guo, Krishna Srinivasan, Ting Chen, Sneha Kudugunta, Chao
  Jia, Yinfei Yang, and Jason Baldridge.
\newblock Mural: multimodal, multitask retrieval across languages.
\newblock {\em arXiv preprint arXiv:2109.05125}, 2021.

\bibitem{jia2021scaling}
Chao Jia, Yinfei Yang, Ye Xia, Yi-Ting Chen, Zarana Parekh, Hieu Pham, Quoc~V
  Le, Yunhsuan Sung, Zhen Li, and Tom Duerig.
\newblock Scaling up visual and vision-language representation learning with
  noisy text supervision.
\newblock {\em arXiv preprint arXiv:2102.05918}, 2021.

\bibitem{johnson2019faiss}
Jeff Johnson, Matthijs Douze, and Herv{\'e} J{\'e}gou.
\newblock Billion-scale similarity search with gpus.
\newblock {\em IEEE Transactions on Big Data}, 7(3):535--547, 2019.

\bibitem{khandelwal2019generalization}
Urvashi Khandelwal, Omer Levy, Dan Jurafsky, Luke Zettlemoyer, and Mike Lewis.
\newblock Generalization through memorization: Nearest neighbor language
  models.
\newblock {\em arXiv preprint arXiv:1911.00172}, 2019.

\bibitem{kingma2014adam}
Diederik~P Kingma and Jimmy Ba.
\newblock Adam: A method for stochastic optimization.
\newblock {\em arXiv preprint arXiv:1412.6980}, 2014.

\bibitem{kolesnikov2020big}
Alexander Kolesnikov, Lucas Beyer, Xiaohua Zhai, Joan Puigcerver, Jessica Yung,
  Sylvain Gelly, and Neil Houlsby.
\newblock Big transfer (bit): General visual representation learning.
\newblock In {\em ECCV}, 2020.

\bibitem{kornblith2019better}
Simon Kornblith, Jonathon Shlens, and Quoc~V Le.
\newblock Do better imagenet models transfer better?
\newblock In {\em Proceedings of the IEEE/CVF conference on computer vision and
  pattern recognition}, pages 2661--2671, 2019.

\bibitem{stanfordcars2013}
Jonathan Krause, Michael Stark, Jia Deng, and Li Fei-Fei.
\newblock 3d object representations for fine-grained categorization.
\newblock In {\em 4th International IEEE Workshop on 3D Representation and
  Recognition (3dRR-13)}, Sydney, Australia, 2013.

\bibitem{lee2022uniclip}
Janghyeon Lee, Jongsuk Kim, Hyounguk Shon, Bumsoo Kim, Seung~Hwan Kim, Honglak
  Lee, and Junmo Kim.
\newblock Uniclip: Unified framework for contrastive language-image
  pre-training.
\newblock {\em arXiv preprint arXiv:2209.13430}, 2022.

\bibitem{lewis2020retrieval}
Patrick Lewis, Ethan Perez, Aleksandra Piktus, Fabio Petroni, Vladimir
  Karpukhin, Naman Goyal, Heinrich K{\"u}ttler, Mike Lewis, Wen-tau Yih, Tim
  Rockt{\"a}schel, et~al.
\newblock Retrieval-augmented generation for knowledge-intensive {NLP} tasks.
\newblock {\em NeurIPS}, 2020.

\bibitem{li2022elevater}
Chunyuan Li, Haotian Liu, Liunian~Harold Li, Pengchuan Zhang, Jyoti Aneja,
  Jianwei Yang, Ping Jin, Houdong Hu, Zicheng Liu, Yong~Jae Lee, and Jianfeng
  Gao.
\newblock {ELEVATER}: A benchmark and toolkit for evaluating language-augmented
  visual models.
\newblock In {\em NeurIPS Track on Datasets and Benchmarks}, 2022.

\bibitem{li2021supervision}
Yangguang Li, Feng Liang, Lichen Zhao, Yufeng Cui, Wanli Ouyang, Jing Shao,
  Fengwei Yu, and Junjie Yan.
\newblock Supervision exists everywhere: A data efficient contrastive
  language-image pre-training paradigm.
\newblock {\em arXiv preprint arXiv:2110.05208}, 2021.

\bibitem{lin2014microsoft}
Tsung-Yi Lin, Michael Maire, Serge Belongie, James Hays, Pietro Perona, Deva
  Ramanan, Piotr Doll{\'a}r, and C~Lawrence Zitnick.
\newblock Microsoft coco: Common objects in context.
\newblock In {\em ECCV}, 2014.

\bibitem{liu2020k}
Weijie Liu, Peng Zhou, Zhe Zhao, Zhiruo Wang, Qi Ju, Haotang Deng, and Ping
  Wang.
\newblock K-{BERT}: Enabling language representation with knowledge graph.
\newblock In {\em AAAI}, 2020.

\bibitem{long2022retrieval}
Alexander Long, Wei Yin, Thalaiyasingam Ajanthan, Vu Nguyen, Pulak Purkait,
  Ravi Garg, Alan Blair, Chunhua Shen, and Anton van~den Hengel.
\newblock Retrieval augmented classification for long-tail visual recognition.
\newblock In {\em Proceedings of the IEEE/CVF Conference on Computer Vision and
  Pattern Recognition}, pages 6959--6969, 2022.

\bibitem{fgvcaircraft2013}
S. Maji, J. Kannala, E. Rahtu, M. Blaschko, and A. Vedaldi.
\newblock Fine-grained visual classification of aircraft.
\newblock Technical report, 2013.

\bibitem{malkov2018hnsw}
Yu~A Malkov and Dmitry~A Yashunin.
\newblock Efficient and robust approximate nearest neighbor search using
  hierarchical navigable small world graphs.
\newblock {\em IEEE transactions on pattern analysis and machine intelligence},
  42(4):824--836, 2018.

\bibitem{marino2021krisp}
Kenneth Marino, Xinlei Chen, Devi Parikh, Abhinav Gupta, and Marcus Rohrbach.
\newblock Krisp: Integrating implicit and symbolic knowledge for open-domain
  knowledge-based {VQA}.
\newblock In {\em CVPR}, 2021.

\bibitem{meyer2012wiktionary}
Christian~M Meyer and Iryna Gurevych.
\newblock {\em Wiktionary: A new rival for expert-built lexicons? Exploring the
  possibilities of collaborative lexicography}.
\newblock na, 2012.

\bibitem{miller1998wordnet}
George~A Miller.
\newblock {\em WordNet: An electronic lexical database}.
\newblock MIT press, 1998.

\bibitem{mishra2021cross}
Swaroop Mishra, Daniel Khashabi, Chitta Baral, and Hannaneh Hajishirzi.
\newblock Cross-task generalization via natural language crowdsourcing
  instructions.
\newblock {\em arXiv preprint arXiv:2104.08773}, 2021.

\bibitem{mu2021slip}
Norman Mu, Alexander Kirillov, David Wagner, and Saining Xie.
\newblock Slip: Self-supervision meets language-image pre-training.
\newblock {\em arXiv preprint arXiv:2112.12750}, 2021.

\bibitem{mustafa2022multimodal}
Basil Mustafa, Carlos Riquelme, Joan Puigcerver, Rodolphe Jenatton, and Neil
  Houlsby.
\newblock Multimodal contrastive learning with limoe: the language-image
  mixture of experts.
\newblock {\em arXiv preprint arXiv:2206.02770}, 2022.

\bibitem{peters2019knowledge}
Matthew~E Peters, Mark Neumann, Robert~L Logan~IV, Roy Schwartz, Vidur Joshi,
  Sameer Singh, and Noah~A Smith.
\newblock Knowledge enhanced contextual word representations.
\newblock {\em arXiv preprint arXiv:1909.04164}, 2019.

\bibitem{pham2021combined}
Hieu Pham, Zihang Dai, Golnaz Ghiasi, Kenji Kawaguchi, Hanxiao Liu, Adams~Wei
  Yu, Jiahui Yu, Yi-Ting Chen, Minh-Thang Luong, Yonghui Wu, et~al.
\newblock Combined scaling for open-vocabulary image classification.
\newblock {\em arXiv preprint arXiv: 2111.10050}, 2021.

\bibitem{plummer2015flickr30k}
Bryan~A Plummer, Liwei Wang, Chris~M Cervantes, Juan~C Caicedo, Julia
  Hockenmaier, and Svetlana Lazebnik.
\newblock Flickr30k entities: Collecting region-to-phrase correspondences for
  richer image-to-sentence models.
\newblock In {\em ICCV}, 2015.

\bibitem{qi2020imagebert}
Di Qi, Lin Su, Jia Song, Edward Cui, Taroon Bharti, and Arun Sacheti.
\newblock Imagebert: Cross-modal pre-training with large-scale weak-supervised
  image-text data.
\newblock {\em arXiv preprint arXiv:2001.07966}, 2020.

\bibitem{radford2021learning}
Alec Radford, Jong~Wook Kim, Chris Hallacy, Aditya Ramesh, Gabriel Goh,
  Sandhini Agarwal, Girish Sastry, Amanda Askell, Pamela Mishkin, Jack Clark,
  et~al.
\newblock Learning transferable visual models from natural language
  supervision.
\newblock {\em arXiv preprint arXiv:2103.00020}, 2021.

\bibitem{saito2022prefix}
Kuniaki Saito, Kihyuk Sohn, Xiang Zhang, Chun-Liang Li, Chen-Yu Lee, Kate
  Saenko, and Tomas Pfister.
\newblock Prefix conditioning unifies language and label supervision.
\newblock {\em arXiv preprint arXiv:2206.01125}, 2022.

\bibitem{schuhmann2022laion}
Christoph Schuhmann, Romain Beaumont, Richard Vencu, Cade Gordon, Ross
  Wightman, Mehdi Cherti, Theo Coombes, Aarush Katta, Clayton Mullis, Mitchell
  Wortsman, et~al.
\newblock Laion-5b: An open large-scale dataset for training next generation
  image-text models.
\newblock {\em arXiv preprint arXiv:2210.08402}, 2022.

\bibitem{schuhmann2021laion}
Christoph Schuhmann, Richard Vencu, Romain Beaumont, Robert Kaczmarczyk,
  Clayton Mullis, Aarush Katta, Theo Coombes, Jenia Jitsev, and Aran
  Komatsuzaki.
\newblock Laion-400m: Open dataset of clip-filtered 400 million image-text
  pairs.
\newblock {\em arXiv preprint arXiv:2111.02114}, 2021.

\bibitem{scudder1965probability}
Henry Scudder.
\newblock Probability of error of some adaptive pattern-recognition machines.
\newblock {\em IEEE Transactions on Information Theory}, 1965.

\bibitem{sharma2018conceptual}
Piyush Sharma, Nan Ding, Sebastian Goodman, and Radu Soricut.
\newblock Conceptual captions: A cleaned, hypernymed, image alt-text dataset
  for automatic image captioning.
\newblock In {\em ACL}, 2018.

\bibitem{shen2022klite}
Sheng Shen, Chunyuan Li, Xiaowei Hu, Yujia Xie, Jianwei Yang, Pengchuan Zhang,
  Anna Rohrbach, Zhe Gan, Lijuan Wang, Lu Yuan, Ce Liu, Kurt Keutzer, Trevor
  Darrell, and Jianfeng Gao.
\newblock {K-LITE}: Learning transferable visual models with external
  knowledge.
\newblock In {\em NeurIPS}, 2022.

\bibitem{sheynin2022knn}
Shelly Sheynin, Oron Ashual, Adam Polyak, Uriel Singer, Oran Gafni, Eliya
  Nachmani, and Yaniv Taigman.
\newblock Knn-diffusion: Image generation via large-scale retrieval.
\newblock {\em arXiv preprint arXiv:2204.02849}, 2022.

\bibitem{sun2017revisiting}
Chen Sun, Abhinav Shrivastava, Saurabh Singh, and Abhinav Gupta.
\newblock Revisiting unreasonable effectiveness of data in deep learning era.
\newblock In {\em ICCV}, 2017.

\bibitem{thomee2016yfcc100m}
Bart Thomee, David~A Shamma, Gerald Friedland, Benjamin Elizalde, Karl Ni,
  Douglas Poland, Damian Borth, and Li-Jia Li.
\newblock Yfcc100m: The new data in multimedia research.
\newblock {\em Communications of the ACM}, 2016.

\bibitem{tsimpoukelli2021multimodal}
Maria Tsimpoukelli, Jacob~L Menick, Serkan Cabi, SM Eslami, Oriol Vinyals, and
  Felix Hill.
\newblock Multimodal few-shot learning with frozen language models.
\newblock {\em Advances in Neural Information Processing Systems}, 2021.

\bibitem{patchcamelyon2018}
Bastiaan~S Veeling, Jasper Linmans, Jim Winkens, Taco Cohen, and Max Welling.
\newblock Rotation equivariant cnns for digital pathology.
\newblock In {\em International Conference on Medical image computing and
  computer-assisted intervention}, pages 210--218. Springer, 2018.

\bibitem{wang2019learning}
Haohan Wang, Songwei Ge, Zachary Lipton, and Eric~P Xing.
\newblock Learning robust global representations by penalizing local predictive
  power.
\newblock In {\em Advances in Neural Information Processing Systems}, pages
  10506--10518, 2019.

\bibitem{wang2022benchmarking}
Yizhong Wang, Swaroop Mishra, Pegah Alipoormolabashi, Yeganeh Kordi, Amirreza
  Mirzaei, Anjana Arunkumar, Arjun Ashok, Arut~Selvan Dhanasekaran, Atharva
  Naik, David Stap, et~al.
\newblock Benchmarking generalization via in-context instructions on 1,600+
  language tasks.
\newblock {\em arXiv preprint arXiv:2204.07705}, 2022.

\bibitem{wu2021multi}
Jialin Wu, Jiasen Lu, Ashish Sabharwal, and Roozbeh Mottaghi.
\newblock Multi-modal answer validation for knowledge-based {VQA}.
\newblock {\em arXiv preprint arXiv:2103.12248}, 2021.

\bibitem{xian2018zero}
Yongqin Xian, Christoph~H Lampert, Bernt Schiele, and Zeynep Akata.
\newblock Zero-shot learning—a comprehensive evaluation of the good, the bad
  and the ugly.
\newblock {\em PAMI}, 2018.

\bibitem{xie2020self}
Qizhe Xie, Minh-Thang Luong, Eduard Hovy, and Quoc~V Le.
\newblock Self-training with noisy student improves imagenet classification.
\newblock In {\em Proceedings of the IEEE/CVF conference on computer vision and
  pattern recognition}, 2020.

\bibitem{yang2022vision}
Jinyu Yang, Jiali Duan, Son Tran, Yi Xu, Sampath Chanda, Liqun Chen, Belinda
  Zeng, Trishul Chilimbi, and Junzhou Huang.
\newblock Vision-language pre-training with triple contrastive learning.
\newblock In {\em Proceedings of the IEEE/CVF Conference on Computer Vision and
  Pattern Recognition}, pages 15671--15680, 2022.

\bibitem{yang2022focal}
Jianwei Yang, Chunyuan Li, and Jianfeng Gao.
\newblock Focal modulation networks.
\newblock {\em Advances in Neural Information Processing Systems}, 2022.

\bibitem{yang2022unicl}
Jianwei Yang, Chunyuan Li, Pengchuan Zhang, Bin Xiao, Lu Yuan, Ce Liu, and
  Jianfeng Gao.
\newblock Unified contrastive learning in image-text-label space.
\newblock {\em CVPR}, 2022.

\bibitem{yang2021empirical}
Zhengyuan Yang, Zhe Gan, Jianfeng Wang, Xiaowei Hu, Yumao Lu, Zicheng Liu, and
  Lijuan Wang.
\newblock An empirical study of {GPT}-3 for few-shot knowledge-based {VQA}.
\newblock {\em arXiv preprint arXiv:2109.05014}, 2021.

\bibitem{yao2021filip}
Lewei Yao, Runhui Huang, Lu Hou, Guansong Lu, Minzhe Niu, Hang Xu, Xiaodan
  Liang, Zhenguo Li, Xin Jiang, and Chunjing Xu.
\newblock Filip: Fine-grained interactive language-image pre-training.
\newblock {\em arXiv preprint arXiv:2111.07783}, 2021.

\bibitem{yasunaga2022retrieval}
Michihiro Yasunaga, Armen Aghajanyan, Weijia Shi, Rich James, Jure Leskovec,
  Percy Liang, Mike Lewis, Luke Zettlemoyer, and Wen-tau Yih.
\newblock Retrieval-augmented multimodal language modeling.
\newblock {\em arXiv preprint arXiv:2211.12561}, 2022.

\bibitem{you2022learning}
Haoxuan You, Luowei Zhou, Bin Xiao, Noel Codella, Yu Cheng, Ruochen Xu, Shih-Fu
  Chang, and Lu Yuan.
\newblock Learning visual representation from modality-shared contrastive
  language-image pre-training.
\newblock In {\em European Conference on Computer Vision}, pages 69--87.
  Springer, 2022.

\bibitem{young2014image}
Peter Young, Alice Lai, Micah Hodosh, and Julia Hockenmaier.
\newblock From image descriptions to visual denotations: New similarity metrics
  for semantic inference over event descriptions.
\newblock {\em Transactions of the Association for Computational Linguistics},
  2:67--78, 2014.

\bibitem{yu2022coca}
Jiahui Yu, Zirui Wang, Vijay Vasudevan, Legg Yeung, Mojtaba Seyedhosseini, and
  Yonghui Wu.
\newblock Coca: Contrastive captioners are image-text foundation models.
\newblock {\em arXiv preprint arXiv:2205.01917}, 2022.

\bibitem{yu2021dict}
Wenhao Yu, Chenguang Zhu, Yuwei Fang, Donghan Yu, Shuohang Wang, Yichong Xu,
  Michael Zeng, and Meng Jiang.
\newblock Dict-bert: Enhancing language model pre-training with dictionary.
\newblock {\em arXiv preprint arXiv:2110.06490}, 2021.

\bibitem{yuan2021florence}
Lu Yuan, Dongdong Chen, Yi-Ling Chen, Noel Codella, Xiyang Dai, Jianfeng Gao,
  Houdong Hu, Xuedong Huang, Boxin Li, Chunyuan Li, et~al.
\newblock Florence: A new foundation model for computer vision.
\newblock {\em arXiv preprint arXiv:2111.11432}, 2021.

\bibitem{zhai2022lit}
Xiaohua Zhai, Xiao Wang, Basil Mustafa, Andreas Steiner, Daniel Keysers,
  Alexander Kolesnikov, and Lucas Beyer.
\newblock Lit: Zero-shot transfer with locked-image text tuning.
\newblock In {\em Proceedings of the IEEE/CVF Conference on Computer Vision and
  Pattern Recognition}, pages 18123--18133, 2022.

\bibitem{zhong2021regionclip}
Yiwu Zhong, Jianwei Yang, Pengchuan Zhang, Chunyuan Li, Noel Codella,
  Liunian~Harold Li, Luowei Zhou, Xiyang Dai, Lu Yuan, Yin Li, et~al.
\newblock Regionclip: Region-based language-image pretraining.
\newblock {\em CVPR}, 2022.

\bibitem{zhou2022maskclip}
Chong Zhou, Chen~Change Loy, and Bo Dai.
\newblock Extract free dense labels from clip.
\newblock In {\em ECCV}, 2022.

\bibitem{zhou2021ibot}
Jinghao Zhou, Chen Wei, Huiyu Wang, Wei Shen, Cihang Xie, Alan Yuille, and Tao
  Kong.
\newblock ibot: Image bert pre-training with online tokenizer.
\newblock {\em arXiv preprint arXiv:2111.07832}, 2021.

\bibitem{zhou2022learning}
Kaiyang Zhou, Jingkang Yang, Chen~Change Loy, and Ziwei Liu.
\newblock Learning to prompt for vision-language models.
\newblock {\em International Journal of Computer Vision}, 130(9):2337--2348,
  2022.

\bibitem{zhou2022lafite2}
Yufan Zhou, Chunyuan Li, Changyou Chen, Jianfeng Gao, and Jinhui Xu.
\newblock Lafite2: Few-shot text-to-image generation.
\newblock {\em arXiv preprint arXiv:2210.14124}, 2022.

\end{thebibliography}
}

\appendix
\section*{Appendix}

This appendix is organized as follows. 

\vspace{6pt}

\begin{itemize}

\item In Section \ref{sec:more_results}, we present more results on ImageNet (Sec.~\ref{sec:more_results_imagenet}) and ELEVATER (Sec.~\ref{sec:more_results_elevater}), with additional studies on a broader selection of checkpoints and more visualizations for a better standing of our approach.

\vspace{-6pt}
\item In Section \ref{sec:implementation_details}, we provide implementation details (Sec.~\ref{sec:details_customization},\ref{sec:retrieval_system}) and cost analysis (Sec.~\ref{sec:details_cost}) of our retrieval system and model customization pipelines.
\end{itemize}

\begin{table*}[t]
    \centering
    \footnotesize
    \scalebox{1.0}{
    \begin{tabular}{c|l|l|lllll}
        \toprule
        $f_{\thetav}$ & Pretrain Data & Method & ImageNet & ImageNet-V2 & ImageNet-R & ImageNet-A & IN-Sketch  \\
        \midrule
\multirow{10}{*}{B/32} & \multirow{3}{*}{WIT-400M} & CLIP & 63.2 & 55.9 & 69.3 & 31.4 & 42.3 \\
 &  & \shortname{} (Locked-Text) & {\bf\cellcolor{Gray}66.9 \textcolor{emerald!80}{(+3.7)}} & {\bf\cellcolor{Gray}58.6 \textcolor{emerald!80}{(+2.7)}} & {\bf\cellcolor{Gray}77.9 \textcolor{emerald!80}{(+8.6)}} & {\bf\cellcolor{Gray}23.0 \textcolor{coralred!80}{(-8.4)}} & {\bf\cellcolor{Gray}54.2 \textcolor{emerald!80}{(+11.8)}} \\
 &  & \shortname{} (Locked-Text Gated-Image) & {\bf\cellcolor{Gray}68.6 \textcolor{emerald!80}{(+5.4)}} & {\bf\cellcolor{Gray}61.0 \textcolor{emerald!80}{(+5.1)}} & {\bf\cellcolor{Gray}78.2 \textcolor{emerald!80}{(+8.9)}} & {\bf\cellcolor{Gray}30.8 \textcolor{coralred!80}{(-0.6)}} & {\bf\cellcolor{Gray}53.9 \textcolor{emerald!80}{(+11.6)}} \\
 \cmidrule{2-8}
 & \multirow{3}{*}{LAION-400M} & OpenCLIP & 62.9 & 55.2 & 73.4 & 21.8 & 49.4 \\
 &  & \shortname{} (Locked-Text) & {\bf\cellcolor{Gray}65.7 \textcolor{emerald!80}{(+2.8)}} & {\bf\cellcolor{Gray}57.3 \textcolor{emerald!80}{(+2.1)}} & {\bf\cellcolor{Gray}77.5 \textcolor{emerald!80}{(+4.1)}} & {\bf\cellcolor{Gray}20.2 \textcolor{coralred!80}{(-1.6)}} & {\bf\cellcolor{Gray}54.8 \textcolor{emerald!80}{(+5.5)}} \\
 &  & \shortname{} (Locked-Text Gated-Image) & {\bf\cellcolor{Gray}66.4 \textcolor{emerald!80}{(+3.5)}} & {\bf\cellcolor{Gray}58.7 \textcolor{emerald!80}{(+3.5)}} & {\bf\cellcolor{Gray}77.8 \textcolor{emerald!80}{(+4.4)}} & {\bf\cellcolor{Gray}22.7 \textcolor{emerald!80}{(+0.9)}} & {\bf\cellcolor{Gray}54.8 \textcolor{emerald!80}{(+5.5)}} \\
 \cmidrule{2-8}
 & \multirow{3}{*}{LAION-2B} & OpenCLIP & 66.6 & 58.2 & 76.5 & 26.2 & 53.5 \\
 &  & \shortname{} (Locked-Text) & {\bf\cellcolor{Gray}67.5 \textcolor{emerald!80}{(+0.9)}} & {\bf\cellcolor{Gray}59.5 \textcolor{emerald!80}{(+1.3)}} & {\bf\cellcolor{Gray}79.1 \textcolor{emerald!80}{(+2.6)}} & {\bf\cellcolor{Gray}23.8 \textcolor{coralred!80}{(-2.5)}} & {\bf\cellcolor{Gray}57.1 \textcolor{emerald!80}{(+3.6)}} \\
 &  & \shortname{} (Locked-Text Gated-Image) & {\bf\cellcolor{Gray}69.5 \textcolor{emerald!80}{(+2.9)}} & {\bf\cellcolor{Gray}61.6 \textcolor{emerald!80}{(+3.5)}} & {\bf\cellcolor{Gray}80.2 \textcolor{emerald!80}{(+3.7)}} & {\bf\cellcolor{Gray}27.9 \textcolor{emerald!80}{(+1.6)}} & {\bf\cellcolor{Gray}58.4 \textcolor{emerald!80}{(+4.8)}} \\
\midrule
\multirow{6.5}{*}{B/16} & \multirow{3}{*}{WIT-400M} & CLIP & 68.6 & 61.8 & 77.6 & 49.7 & 48.3 \\
 &  & \shortname{} (Locked-Text) & {\bf\cellcolor{Gray}71.6 \textcolor{emerald!80}{(+3.0)}} & {\bf\cellcolor{Gray}64.4 \textcolor{emerald!80}{(+2.6)}} & {\bf\cellcolor{Gray}83.4 \textcolor{emerald!80}{(+5.8)}} & {\bf\cellcolor{Gray}38.8 \textcolor{coralred!80}{(-10.9)}} & {\bf\cellcolor{Gray}58.3 \textcolor{emerald!80}{(+10.0)}} \\
 &  & \shortname{} (Locked-Text Gated-Image) & {\bf\cellcolor{Gray}73.4 \textcolor{emerald!80}{(+4.8)}} & {\bf\cellcolor{Gray}66.8 \textcolor{emerald!80}{(+5.0)}} & {\bf\cellcolor{Gray}84.0 \textcolor{emerald!80}{(+6.4)}} & {\bf\cellcolor{Gray}48.5 \textcolor{coralred!80}{(-1.2)}} & {\bf\cellcolor{Gray}58.3 \textcolor{emerald!80}{(+10.1)}} \\
 \cmidrule{2-8}
 & \multirow{3}{*}{LAION-400M} & OpenCLIP & 67.1 & 59.4 & 77.9 & 33.0 & 52.4 \\
 &  & \shortname{} (Locked-Text) & {\bf\cellcolor{Gray}69.9 \textcolor{emerald!80}{(+2.8)}} & {\bf\cellcolor{Gray}62.4 \textcolor{emerald!80}{(+3.0)}} & {\bf\cellcolor{Gray}81.8 \textcolor{emerald!80}{(+3.9)}} & {\bf\cellcolor{Gray}33.7 \textcolor{emerald!80}{(+0.7)}} & {\bf\cellcolor{Gray}58.1 \textcolor{emerald!80}{(+5.7)}} \\
 &  & \shortname{} (Locked-Text Gated-Image) & {\bf\cellcolor{Gray}70.5 \textcolor{emerald!80}{(+3.4)}} & {\bf\cellcolor{Gray}63.0 \textcolor{emerald!80}{(+3.6)}} & {\bf\cellcolor{Gray}82.3 \textcolor{emerald!80}{(+4.3)}} & {\bf\cellcolor{Gray}37.8 \textcolor{emerald!80}{(+4.9)}} & {\bf\cellcolor{Gray}57.4 \textcolor{emerald!80}{(+5.1)}} \\
\midrule
\multirow{4.5}{*}{L/14} & \multirow{2}{*}{WIT-400M} & CLIP & 75.3 & 69.6 & 87.8 & 70.5 & 59.6 \\
 &  & \shortname{} (Locked-Text Gated-Image) & {\bf\cellcolor{Gray}78.1 \textcolor{emerald!80}{(+2.8)}} & {\bf\cellcolor{Gray}71.5 \textcolor{emerald!80}{(+1.9)}} & {\bf\cellcolor{Gray}89.9 \textcolor{emerald!80}{(+2.1)}} & {\bf\cellcolor{Gray}68.6 \textcolor{coralred!80}{(-2.0)}} & {\bf\cellcolor{Gray}64.8 \textcolor{emerald!80}{(+5.2)}} \\
 \cmidrule{2-8}
 & \multirow{2}{*}{LAION-2B} & OpenCLIP & 75.3 & 67.9 & 84.1 & 42.0 & 63.3 \\
 &  & \shortname{} (Locked-Text Gated-Image) & {\bf\cellcolor{Gray}76.4 \textcolor{emerald!80}{(+1.1)}} & {\bf\cellcolor{Gray}68.9 \textcolor{emerald!80}{(+1.0)}} & {\bf\cellcolor{Gray}89.0 \textcolor{emerald!80}{(+4.9)}} & {\bf\cellcolor{Gray}55.2 \textcolor{emerald!80}{(+13.2)}} & {\bf\cellcolor{Gray}65.4 \textcolor{emerald!80}{(+2.0)}} \\
        \bottomrule
    \end{tabular}
    }
    \caption{Comparison with public checkpoints from CLIP~\cite{radford2021learning} and OpenCLIP~\cite{openclip}.  All \shortname{} checkpoints use 10M retrieved samples from LAION-400M~\cite{schuhmann2021laion} dataset during model customization stage.  It consistently outperforms base CLIP and OpenCLIP checkpoints.}
    \label{tab:more_perf_in1k_zeroshot}
\end{table*}

\section{More Results on Image-level Tasks}
\label{sec:more_results}

\subsection{ImageNet}

\label{sec:more_results_imagenet}

\paragraph{Comparison of a broader selection of checkpoints.}
To further study the improvement of \shortname{} over the pretrained vision-language models, we present more results with different pretraining data (WIT-400M, LAION-400M, LAION-2B) and different vision Transformer model sizes (B/32, B/16, L/14).  For a fair study, we use the same set of 10M retrieved image-text pairs from LAION-400M dataset for all configurations.  Results are presented in Table~\ref{tab:more_perf_in1k_zeroshot} (first column).

From the table, we can see that \shortname{} with both tuning strategies consistently improves over the base checkpoints across different pretrained data and different vision backbones, and locked-text-gated-image tuning consistently performs better than the locked-text tuning only.
Though both benefiting from \shortname{} customization, CLIP checkpoints that are trained on WIT-400M data benefit slightly more than OpenCLIP checkpoints.  This suggests that during the model customization stage, leveraging unseen data can potentially give the model a larger gain compared with the seen data during pretraining.

We further study the case when all retrieval data is already observed by models during their pretraining stage.  Specifically, we study OpenCLIP checkpoints pretrained on LAION-400M~\cite{schuhmann2021laion}, and LAION-2B~\cite{schuhmann2022laion} (a super-set of LAION-400M).
From the results, we see that by revisiting the already observed LAION-400M data, \shortname{} (locked-text) shows +2.8/+3.0 improvements on B32/B16 checkpoints, respectively, which purely comes from the model customization stage, with neither additional model parameters, nor additional training data.  Interestingly, even on OpenCLIP checkpoints that is pretrained with a much larger LAION-2B, \shortname{} can still improves over OpenCLIP by +0.9/+2.9 with B32 backbone with locked-text and locked-text-gated-image tuning strategy, respectively.
These findings suggest that leveraging the original pretraining dataset only at the pre-training stage is sub-optimal, it is of much larger potential to explore the web-scale data using the proposed model customization stage. 

\paragraph{Robustness.}
We also conduct zero-shot evaluation on other ImageNet variants: ImageNet-V2~\cite{kornblith2019better}, ImageNet-R~\cite{hendrycks2021many}, ImageNet-A~\cite{hendrycks2021nae}, ImageNet-Sketch~\cite{wang2019learning} in Table~\ref{tab:more_perf_in1k_zeroshot}.
With the \shortname{} customization, model robustness towards different ImageNet variants consistently improves on ImageNet-V2, ImageNet-R, and ImageNet-Sketch.  We notice that for some checkpoints, the accuracy drops after model customization on ImageNet-R dataset: an adversarial dataset with a collection of selected images from the web that can ``fool'' common classifiers.  We find that classifiers trained on the LAION dataset are more prone to such adversarial attacks, while \shortname{} customization helps it recover from such attacks to some extent: accuracy improves for OpenCLIP checkpoints that are trained on LAION, especially when locked-text gated-image strategy is used.

\paragraph{Linear Probe.}
We further study the full-shot performance on ImageNet-1K of \shortname{} using the linear probing protocol.  ImageNet-1K contains around 1.28M training images, and it represents one of the most standard data-rich settings.  We use the DINO~\cite{caron2021emerging} code base for the linear probe experiments.  As shown in Table~\ref{tab:linear_probe_in1k}, \shortname{} improves over CLIP by +0.6/+1.9 with the locked-text and locked-text-gated-image tuning, respectively.  This suggests that the \shortname{} customization adequately adapts the visual encoder to the ImageNet domain, resulting in better feature representations.

\begin{table}[t]
    \centering
    \footnotesize
    \begin{tabular}{l|l}
        \toprule
        Method & Accuracy \\
        \midrule
        CLIP~\cite{radford2021learning} & 80.2 \\
        \midrule
        CLIP$^\dagger$ & 79.5 \\
        \rowcolor{Gray} \shortname{} (Locked-Text) & {\bf 80.1 \textcolor{emerald!80}{(+0.6)}} \\
        \rowcolor{Gray} \shortname{} (Locked-Text Gated-Image) & {\bf 81.4 \textcolor{emerald!80}{(+1.9)}} \\
        \bottomrule
    \end{tabular}
    \caption{Linear Probe on ImageNet-1K. CLIP$^\dagger$: reproduced by our implementation.
    }
    \label{tab:linear_probe_in1k}
\end{table}

\paragraph{Sample Overlap.}
There is a chance that the LAION-400M dataset contains \emph{some} of downstream ImageNet images, and our retrieval system \emph{may} retrieve these image-text pairs.  One may question that if the performance gain of \shortname{} model customization actually comes from these samples.

\begin{table}[t]
    \centering
    \footnotesize
    \begin{tabular}{l|cc}
        \toprule
        & \multicolumn{2}{c}{Accuracy} \\
        Method & Filtered & Unfiltered \\
        \midrule
        OpenCLIP~\cite{openclip} & -- & 62.9 \\
        \midrule
        \rowcolor{Gray} \shortname{} (Locked-Text) & 66.7 & 66.9 \\
        \rowcolor{Gray} \shortname{} (Locked-Text Gated-Image) & 68.4 & 68.6 \\
        \bottomrule
    \end{tabular}
    \caption{The study of sample overlap. Comparison between checkpoints trained on filtered and unfiltered retrieved set.  \shortname{} is customized based on CLIP ViT-B/32 checkpoint, whose ImageNet zeroshot accuracy is 63.2.
    \label{tab:compare_filter}
    }
\end{table}

\begin{figure}[t]
	\centering
	\includegraphics[width=\linewidth]{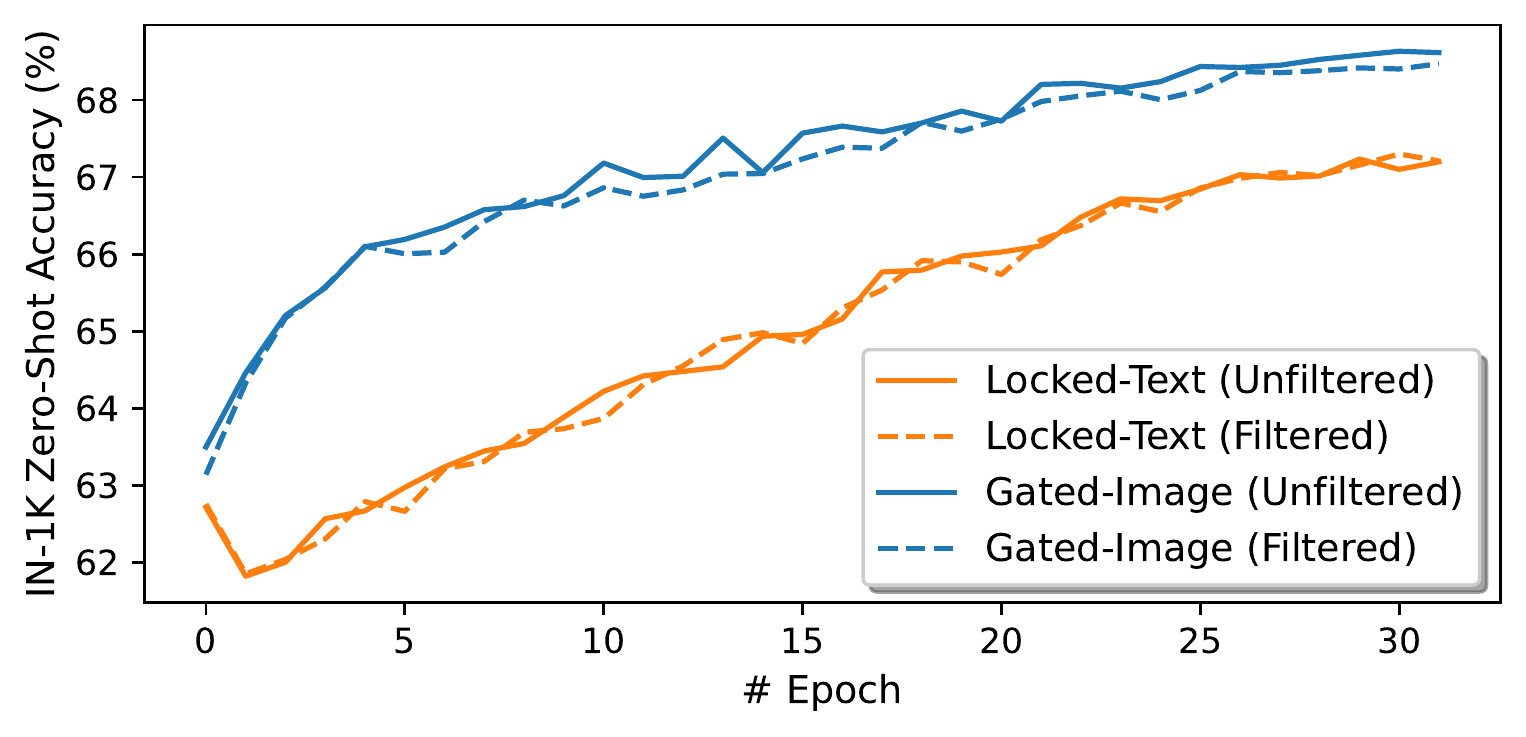}
	\includegraphics[width=\linewidth]{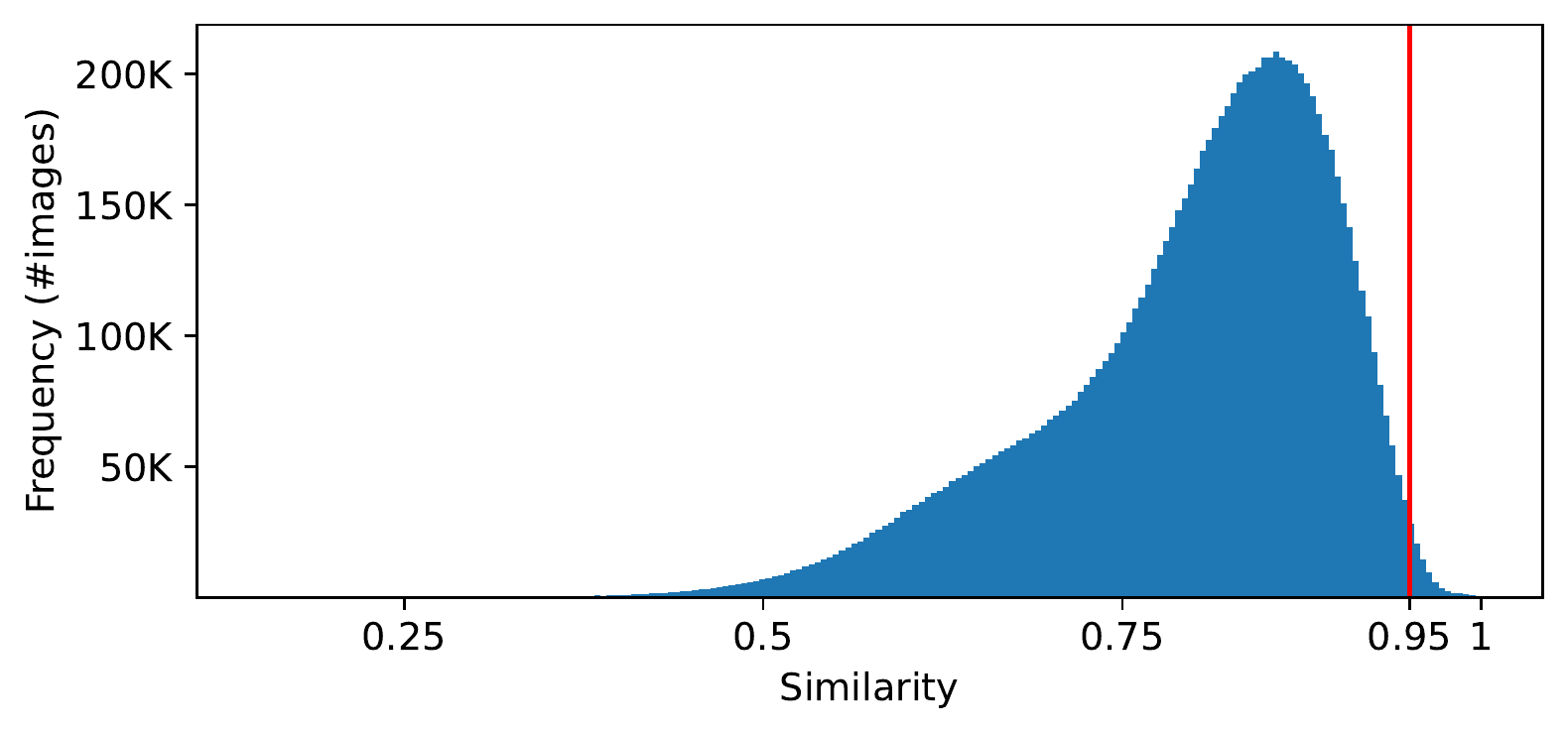}
    \caption{The study of sample overlap.  Top: comparison of validation accuracy curve between checkpoints using filtered (dashed line) and unfiltered (solid line) retrieval set during the model customization stage.  Both the training behavior and the final model performance are similar.  Bottom: histogram of the sample similarity between the retrieved samples and ImageNet training set (nbins=200). The cutoff threshold is set to 0.95 (red line). }
    \label{fig:compare_filter_training_curve}
\end{figure}

\begin{table*}[ht!]
    \centering
    \footnotesize
    \addtolength{\tabcolsep}{-3.5pt}
    \scalebox{0.96}{
    \begin{tabular}{l|c|cccccccccccccccccccc}
        \toprule
        Strategy & Score & \rotatebox[origin=l]{90}{Caltech101} & \rotatebox[origin=l]{90}{CIFAR10} & \rotatebox[origin=l]{90}{CIFAR100} & \rotatebox[origin=l]{90}{Country211} & \rotatebox[origin=l]{90}{DTD} & \rotatebox[origin=l]{90}{EuroSat} & \rotatebox[origin=l]{90}{FER2013} & \rotatebox[origin=l]{90}{FGVCAircraft} & \rotatebox[origin=l]{90}{Food101} & \rotatebox[origin=l]{90}{GTSRB} & \rotatebox[origin=l]{90}{HatefulMemes} & \rotatebox[origin=l]{90}{KittiDistance} & \rotatebox[origin=l]{90}{MNIST} & \rotatebox[origin=l]{90}{Flowers102} & \rotatebox[origin=l]{90}{OxfordPets} & \rotatebox[origin=l]{90}{PatchCamelyon} & \rotatebox[origin=l]{90}{SST2} & \rotatebox[origin=l]{90}{RESISC45} & \rotatebox[origin=l]{90}{StanfordCars} & \rotatebox[origin=l]{90}{VOC2007} \\
        \midrule
\multicolumn{22}{l}{ \bf{Zero-Shot Adaptation}} \\
 & 56.8 & 87.5 & 89.9 & 65.1 & 17.2 & 44.4 & 45.6 & 42.1 & 19.6 & 84.0 & 32.8 & 56.0 & 29.0 & 48.1 & 66.5 & 87.1 & 60.7 & 58.4 & 60.0 & 59.7 & 82.6 \\
Locked-Text & {\bf\cellcolor{emerald!30}59.3} & {\bf\cellcolor{emerald!30}90.7} & {\bf\cellcolor{emerald!30}94.0} & {\bf\cellcolor{emerald!30}73.7} & 17.1 & {\bf\cellcolor{emerald!30}47.6} & {\bf\cellcolor{emerald!30}53.4} & {\bf\cellcolor{emerald!30}46.8} & {\bf\cellcolor{emerald!30}28.1} & 83.0 & 27.2 & 54.8 & 25.2 & {\bf\cellcolor{emerald!30}50.6} & {\bf\cellcolor{emerald!30}74.0} & {\bf\cellcolor{emerald!30}88.3} & 48.1 & 54.5 & 59.7 & {\bf\cellcolor{emerald!30}87.3} & 81.0 \\
Locked-Text Gated-Image & {\bf\cellcolor{emerald!30}60.7} & {\bf\cellcolor{emerald!30}90.6} & {\bf\cellcolor{emerald!30}91.7} & {\bf\cellcolor{emerald!30}70.7} & {\bf\cellcolor{emerald!30}19.1} & {\bf\cellcolor{emerald!30}49.2} & {\bf\cellcolor{emerald!30}53.8} & {\bf\cellcolor{emerald!30}49.0} & {\bf\cellcolor{emerald!30}30.0} & {\bf\cellcolor{emerald!30}85.1} & 30.5 & 54.1 & 22.2 & {\bf\cellcolor{emerald!30}53.9} & {\bf\cellcolor{emerald!30}76.1} & {\bf\cellcolor{emerald!30}90.1} & 53.9 & {\bf\cellcolor{emerald!30}58.4} & {\bf\cellcolor{emerald!30}62.9} & {\bf\cellcolor{emerald!30}88.9} & {\bf\cellcolor{emerald!30}82.8} \\

\midrule
\multicolumn{22}{l}{ \bf{Few-Shot Linear Probe}} \\
 & 65.3 & 89.8 & 90.0 & 67.4 & 17.5 & 59.6 & 73.2 & 47.4 & 28.4 & 84.2 & 52.5 & 56.0 & 44.9 & 71.1 & 90.5 & 88.0 & 63.2 & 57.5 & 76.6 & 65.0 & 84.0 \\
Locked-Text & {\bf\cellcolor{emerald!30}69.4} & {\bf\cellcolor{emerald!30}92.2} & {\bf\cellcolor{emerald!30}94.1} & {\bf\cellcolor{emerald!30}76.3} & {\bf\cellcolor{emerald!30}17.6} & {\bf\cellcolor{emerald!30}66.6} & {\bf\cellcolor{emerald!30}82.5} & {\bf\cellcolor{emerald!30}49.9} & {\bf\cellcolor{emerald!30}42.1} & {\bf\cellcolor{emerald!30}84.2} & {\bf\cellcolor{emerald!30}55.0} & 54.9 & 42.2 & {\bf\cellcolor{emerald!30}78.7} & {\bf\cellcolor{emerald!30}96.7} & {\bf\cellcolor{emerald!30}89.0} & 58.5 & 54.2 & {\bf\cellcolor{emerald!30}80.5} & {\bf\cellcolor{emerald!30}89.3} & 83.3 \\
Locked-Text Gated-Image & {\bf\cellcolor{emerald!30}68.9} & {\bf\cellcolor{emerald!30}92.5} & {\bf\cellcolor{emerald!30}92.3} & {\bf\cellcolor{emerald!30}71.6} & {\bf\cellcolor{emerald!30}18.9} & {\bf\cellcolor{emerald!30}66.2} & {\bf\cellcolor{emerald!30}74.5} & {\bf\cellcolor{emerald!30}51.8} & {\bf\cellcolor{emerald!30}44.1} & {\bf\cellcolor{emerald!30}85.6} & 51.9 & 54.1 & 42.9 & 68.7 & {\bf\cellcolor{emerald!30}97.0} & {\bf\cellcolor{emerald!30}90.6} & 60.5 & {\bf\cellcolor{emerald!30}60.7} & {\bf\cellcolor{emerald!30}78.8} & {\bf\cellcolor{emerald!30}90.0} & {\bf\cellcolor{emerald!30}84.4} \\

\midrule
\multicolumn{22}{l}{ \bf{Few-Shot Full-Model Finetune}} \\
 & 63.3 & 88.8 & 91.3 & 73.0 & 16.6 & 51.8 & 79.3 & 52.3 & 23.1 & 84.0 & 60.4 & 55.8 & 44.3 & 60.5 & 67.3 & 86.9 & 61.8 & 59.3 & 70.8 & 56.3 & 82.4 \\
Locked-Text & {\bf\cellcolor{emerald!30}68.8} & {\bf\cellcolor{emerald!30}93.4} & {\bf\cellcolor{emerald!30}94.2} & {\bf\cellcolor{emerald!30}79.4} & {\bf\cellcolor{emerald!30}16.9} & {\bf\cellcolor{emerald!30}61.2} & 76.0 & 52.2 & {\bf\cellcolor{emerald!30}41.1} & 83.2 & {\bf\cellcolor{emerald!30}77.3} & 54.9 & 44.0 & {\bf\cellcolor{emerald!30}67.5} & {\bf\cellcolor{emerald!30}90.0} & {\bf\cellcolor{emerald!30}88.9} & 57.8 & 53.3 & {\bf\cellcolor{emerald!30}78.0} & {\bf\cellcolor{emerald!30}89.4} & 77.9 \\
Locked-Text Gated-Image & {\bf\cellcolor{emerald!30}68.4} & {\bf\cellcolor{emerald!30}91.3} & {\bf\cellcolor{emerald!30}92.2} & {\bf\cellcolor{emerald!30}77.2} & {\bf\cellcolor{emerald!30}18.1} & {\bf\cellcolor{emerald!30}60.1} & {\bf\cellcolor{emerald!30}81.2} & {\bf\cellcolor{emerald!30}52.6} & {\bf\cellcolor{emerald!30}31.8} & {\bf\cellcolor{emerald!30}85.4} & {\bf\cellcolor{emerald!30}69.4} & 54.1 & 40.0 & {\bf\cellcolor{emerald!30}68.3} & {\bf\cellcolor{emerald!30}88.8} & {\bf\cellcolor{emerald!30}89.7} & 61.0 & {\bf\cellcolor{emerald!30}59.9} & {\bf\cellcolor{emerald!30}77.2} & {\bf\cellcolor{emerald!30}87.0} & {\bf\cellcolor{emerald!30}83.3} \\

\midrule
\multicolumn{22}{l}{ \bf{Full-Shot Linear Probe}} \\
 & 78.4 & 86.0 & 95.1 & 79.8 & 25.9 & 75.3 & 93.8 & 67.8 & 44.7 & 88.6 & 86.9 & 63.1 & 65.8 & 98.8 & 94.5 & 91.0 & 83.2 & 71.6 & 88.1 & 82.1 & 86.0 \\
Locked-Text & {\bf\cellcolor{emerald!30}80.1} & {\bf\cellcolor{emerald!30}94.5} & {\bf\cellcolor{emerald!30}96.6} & {\bf\cellcolor{emerald!30}84.1} & 24.2 & {\bf\cellcolor{emerald!30}77.4} & {\bf\cellcolor{emerald!30}95.7} & 66.0 & {\bf\cellcolor{emerald!30}57.1} & 88.0 & 86.4 & 59.7 & {\bf\cellcolor{emerald!30}68.1} & 98.6 & {\bf\cellcolor{emerald!30}98.1} & {\bf\cellcolor{emerald!30}92.5} & {\bf\cellcolor{emerald!30}83.4} & 63.5 & {\bf\cellcolor{emerald!30}89.4} & {\bf\cellcolor{emerald!30}93.1} & 85.0 \\
Locked-Text Gated-Image & {\bf\cellcolor{emerald!30}80.4} & {\bf\cellcolor{emerald!30}94.5} & {\bf\cellcolor{emerald!30}95.6} & {\bf\cellcolor{emerald!30}81.6} & {\bf\cellcolor{emerald!30}26.3} & {\bf\cellcolor{emerald!30}77.8} & {\bf\cellcolor{emerald!30}95.3} & 67.2 & {\bf\cellcolor{emerald!30}56.5} & {\bf\cellcolor{emerald!30}89.2} & 83.3 & 62.6 & 65.3 & 98.4 & {\bf\cellcolor{emerald!30}98.2} & {\bf\cellcolor{emerald!30}93.6} & {\bf\cellcolor{emerald!30}83.3} & 70.4 & {\bf\cellcolor{emerald!30}89.7} & {\bf\cellcolor{emerald!30}93.5} & {\bf\cellcolor{emerald!30}86.0} \\

\midrule
\multicolumn{22}{l}{ \bf{Full-Shot Full-Model Finetune}} \\
 & 80.3 & 94.0 & 97.8 & 87.0 & 19.1 & 70.1 & 98.1 & 68.9 & 50.7 & 87.7 & 98.6 & 61.9 & 81.0 & 99.5 & 88.5 & 91.6 & 91.0 & 70.6 & 89.4 & 75.8 & 85.7 \\
Locked-Text & {\bf\cellcolor{emerald!30}82.2} & {\bf\cellcolor{emerald!30}95.3} & {\bf\cellcolor{emerald!30}98.3} & {\bf\cellcolor{emerald!30}89.0} & {\bf\cellcolor{emerald!30}20.6} & {\bf\cellcolor{emerald!30}75.1} & 98.0 & {\bf\cellcolor{emerald!30}71.6} & {\bf\cellcolor{emerald!30}60.2} & {\bf\cellcolor{emerald!30}88.0} & {\bf\cellcolor{emerald!30}98.7} & 58.1 & 79.2 & {\bf\cellcolor{emerald!30}99.7} & {\bf\cellcolor{emerald!30}95.3} & {\bf\cellcolor{emerald!30}93.4} & 90.4 & 65.1 & {\bf\cellcolor{emerald!30}90.2} & {\bf\cellcolor{emerald!30}92.6} & 85.1 \\
Locked-Text Gated-Image & {\bf\cellcolor{emerald!30}81.8} & {\bf\cellcolor{emerald!30}94.6} & {\bf\cellcolor{emerald!30}98.3} & {\bf\cellcolor{emerald!30}87.8} & {\bf\cellcolor{emerald!30}19.5} & {\bf\cellcolor{emerald!30}72.5} & 97.9 & {\bf\cellcolor{emerald!30}70.5} & {\bf\cellcolor{emerald!30}59.7} & {\bf\cellcolor{emerald!30}88.4} & {\bf\cellcolor{emerald!30}98.7} & 58.5 & 73.4 & {\bf\cellcolor{emerald!30}99.6} & {\bf\cellcolor{emerald!30}94.5} & {\bf\cellcolor{emerald!30}93.0} & 89.1 & 70.5 & {\bf\cellcolor{emerald!30}89.7} & {\bf\cellcolor{emerald!30}93.0} & {\bf\cellcolor{emerald!30}86.6} \\
        \bottomrule
    \end{tabular}
    }
    \addtolength{\tabcolsep}{3.5pt}
    \vspace{3mm}
    \caption{Full-spectrum breakdown results on \textsc{Elevater} using CLIP (ViT-B/32) and 10M retrieved image-text pairs from LAION-400M.
    }
    \label{table:experiment_breakdown}
\end{table*}

We carefully study the de-duplication experiments. We compute the pairwise distance of the visual features between the images from the retrieved set and the ImageNet train/val set, and set the cutoff threshold to 0.95 (Fig.~\ref{fig:compare_filter_training_curve}, Bottom). Note that 0.95 is a high threshold, as $\sim$85K ($\sim$1\% of 10M total retrieved images) images are removed, among which, only a few of them overlap with ImageNet train/val. %
This suggests that the LAION data contain ImageNet images, making the publicly available OpenCLIP checkpoints less rigorous when reporting the zero-shot task transfer performance. As for CLIP, as its pre-training data is not publicly available, it remains unknown if any ImageNet images are observed in its pre-training. 
We set it to 0.95 mainly to ensure that the overlapping images are removed from the retrieved sets so as to carefully study its effect.

As shown in Table~\ref{tab:compare_filter}, even after aggressively removing 85K images, the final model's performance is similar (-0.2\%) to the checkpoint trained on the unfiltered retrieved set.  We further visualize the validation accuracy curve as training proceeds in Fig.~\ref{fig:compare_filter_training_curve} (Top). The model behaviors during training are very similar between filtered (dashed curve) and unfiltered (solid curve) retrieved sets, for both tuning strategies.
This ensures that the gains are not due to the overlapping samples, and \shortname{} effectively learns and adapts to the ImageNet domain during the model customization stage.

\begin{figure}[t]
	\centering
	\includegraphics[width=\linewidth]{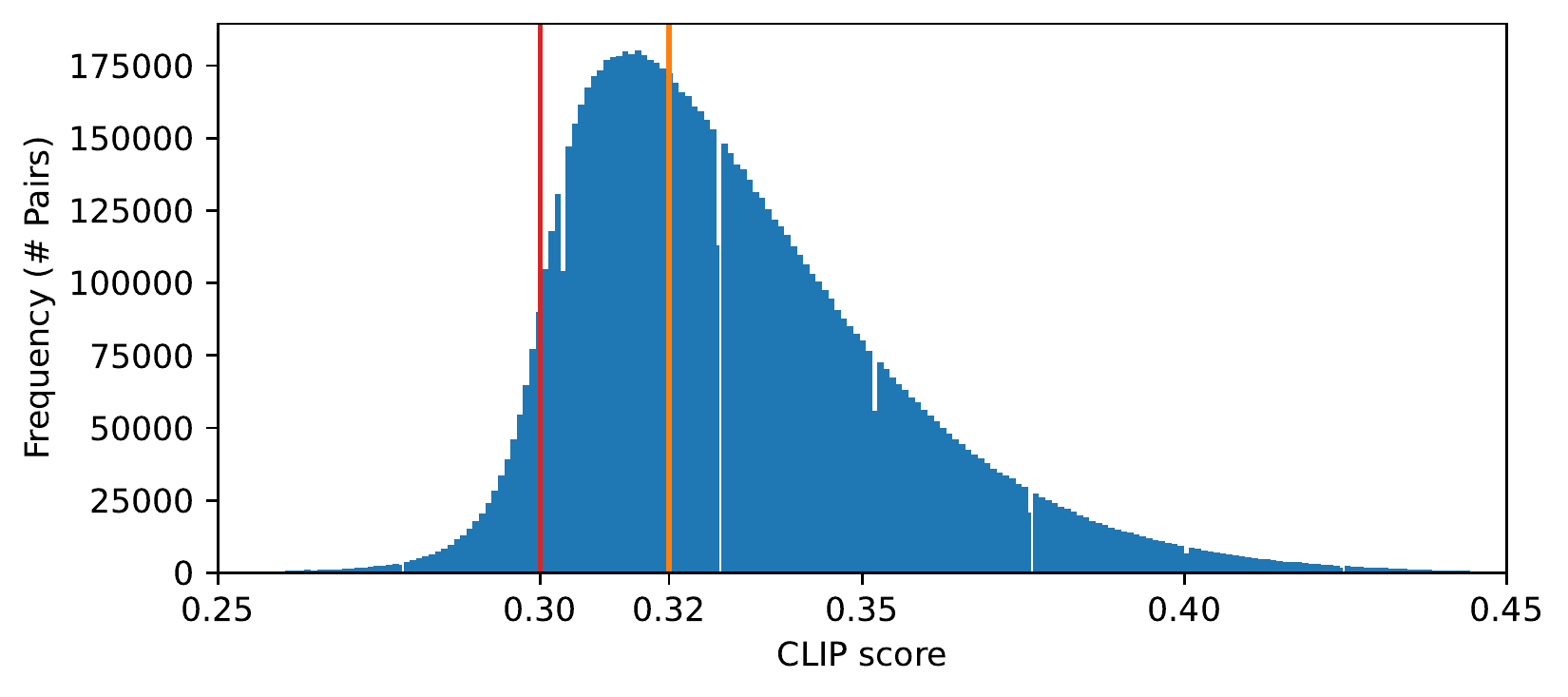}
    \caption{Quality control.  Sample frequency under different CLIP scores.}
    \label{fig:filter_clip_score}
\end{figure}

\begin{table}[t]
    \centering
    \footnotesize
    \scalebox{0.9}{
    \begin{tabular}{l|ccc}
        \toprule
        & \multicolumn{3}{c}{Accuracy} \\
        Method & $\gamma=0.32$ & $\gamma=0.30$ & Unfiltered \\
        \midrule
        CLIP~\cite{radford2021learning} & -- & -- & 63.2 \\
        \midrule
        \rowcolor{Gray} \shortname{} (Locked-Text) & 65.2 & 67.1 & 66.9 \\
        \rowcolor{Gray} \shortname{} (Locked-Text Gated-Image) & 67.3 & 68.1 & 68.6 \\
        \bottomrule
    \end{tabular}
    }
    \caption{Quality control. Comparison between checkpoints trained on CLIP-score-filtered and unfiltered retrieved set.  \shortname{} is customized based on CLIP ViT-B/32 checkpoint.
    \label{tab:quality_control}
    }
\end{table}

\paragraph{Quality Control.}
The quality of retrieved data matters for vision-language pre-training.  As an initial attempt of the quality control, we consider to use the CLIP score to select the high relevant retrieved image-text pairs.  The distribution of the CLIP score is visualized at Fig.~\ref{fig:filter_clip_score}.  We choose CLIP score 0.3 and 0.32 as two thresholds ($\gamma$): a threshold of 0.3 filters the low-quality samples while keeping the total number of retrieved samples roughly the same (93.5\% retrieved samples are kept), a threshold of 0.32 performs a more aggressive filtering and keeps around 6M samples (sufficient for \shortname{} customization according to Sec.~\ref{sec:ablation}).

As shown in Table~\ref{tab:quality_control}, \shortname{} is robust towards noise in the pretraining data. When filtering using a CLIP score threshold of 0.3, the model customization performance roughly remains the same.  When filtering with a threshold of 0.32, the customization performance drops by around 1\%, which suggests that the filtered samples contain useful information for model customization.
In conclusion, our \shortname{} model customization is robust against the noises in the retrieval dataset.  Therefore, we leave a more sophisticated quality control approach to future work.

\subsection{ELEVATER}

\label{sec:more_results_elevater}

\paragraph{Breakdown results.}
We present the full-spectrum breakdown results for zero-shot, few-shot, and full-shot experiments on the ELEVATER benchmark in Table~\ref{table:experiment_breakdown}.  We mark the experiment runs that \shortname{} yields gains compared with baseline CLIP in green and with bold font.

First, across zero-shot, few-shot, and full-shot settings, \shortname{} consistently improves over the baseline CLIP.  However, on the ELEVATER benchmark, locked-text and locked-text-gated-image strategy works better in different cases: for zero-shot, locked-text-gated-image works much better than locked-text with 1.4\% improvement; while for other cases, they perform similarly well, and locked-text is slightly better in 3/4 cases.  This may be partly due to that we are training a unified checkpoint across different domains, and random noises during the final adaptation stage can cause small variations on different datasets.

Second, we find that across different data regimes and different tuning strategies, the gains and losses are mostly consistent on a fixed set of datasets.  This provides another clue for that the gains and losses are highly correlated with the retrieval quality, and the gain/loss conclusion can generally transfer to different tuning configurations.

\begin{figure}[t]
	\centering
	\includegraphics[width=\linewidth]{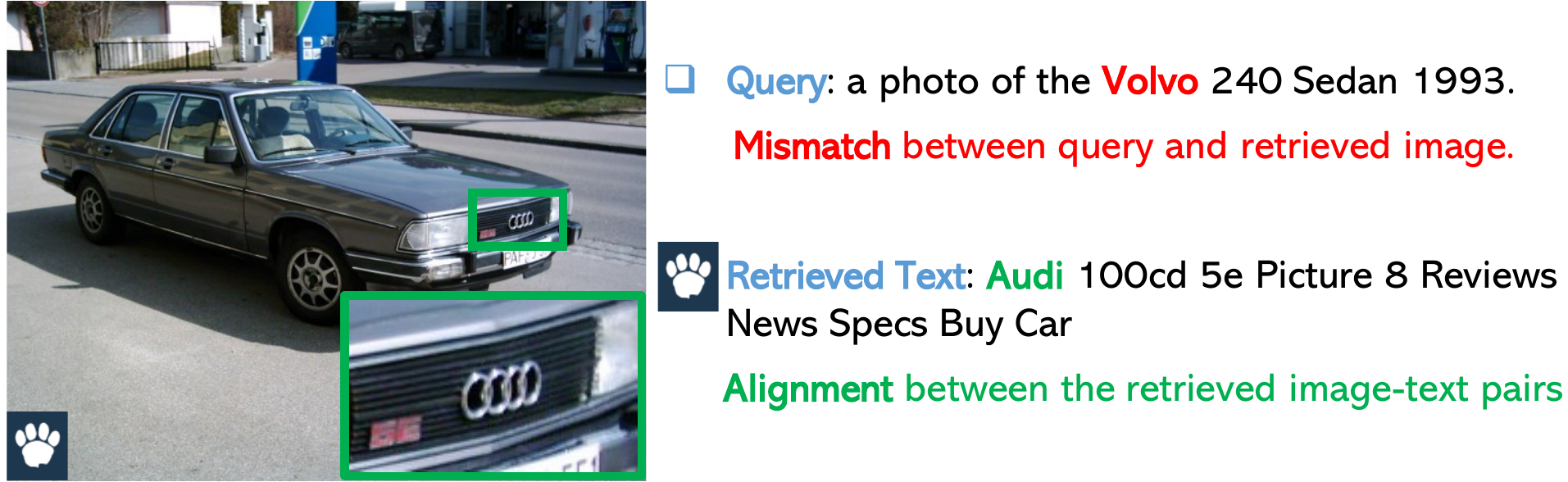}
    \caption{Mismatch between the query (Volvo) and the retrieved image (Audi).  The retrieved caption helps to correct the mistake.}
    \label{fig:example_retrieve_correct_mistake}
\end{figure}

\paragraph{Visualizations.}
We present more retrieved samples from the ELEVATER datasets to illustrate properties of \shortname{}.

First, we show one example of the benefit of using retrieved image-text pairs for training instead of using pseudo labels.  As shown in Fig.~\ref{fig:example_retrieve_correct_mistake}, the query is \texttt{Volvo} sedan, while one of the retrieved sample is an \texttt{Audi}.  The retrieved text contains the correct brand \texttt{Audi}, and the alignment between the retrieved image-text pairs can help correct the retrieval mistake and aid the model training. If the retrieved images are annotated as the same label with query, and used the pseudo-labelled pairs for training, the aforementioned finding suggests that the approach would perform worse than leverage the ``true'' image-text pair knowledge crawled from the web. 

\begin{figure}[t]
	\centering
	\includegraphics[width=\linewidth]{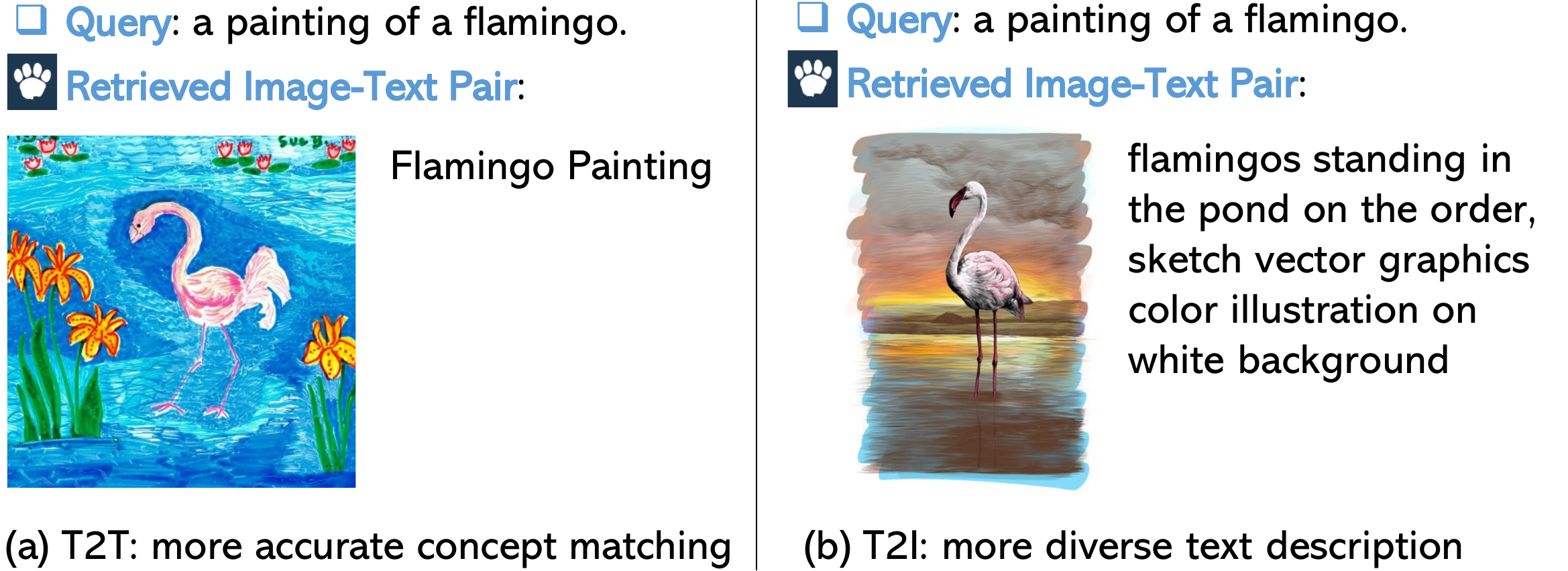}
    \caption{Comparison between T2T and T2I retrieved samples.}
    \label{fig:example_t2t_t2i_comp}
\end{figure}

Second, we visualize examples retrieved by text-to-text and text-to-image retrieval, using the same query: ``\texttt{a painting of a flamingo}''.  As shown in Fig.~\ref{fig:example_t2t_t2i_comp}, text-to-text retrieval can retrieve samples that is more accurately matching the query (``\texttt{a painting}''). On the other hand, text-to-image retrieval gives a more diverse text description, while it may not have a perfect match between the query and the retrieved text sample (``\texttt{sketch vector}'').

\section{Implementation details}
\label{sec:implementation_details}

\subsection{Training Details in Customization}

\label{sec:details_customization}

\paragraph{Model architecture.}
We mainly conduct our experiments on the vision Transformer backbones.  For the ViT architecture, we mainly follow the implementation from CLIP~\cite{radford2021learning}.  The feature from the \texttt{CLS} token from the last visual encoder layer is used as the visual feature.  For the gated-image experiments, we only add gated blocks to the last 6 layers.  The hidden/embedding dimensions for gated blocks are set the same as the layer that it is added to.  Following \cite{alayrac2022flamingo}, the gate values are initialized as zero, modulated by \texttt{tanh} operator.

\paragraph{Training Protocol.}
We mainly follow CLIP~\cite{radford2021learning} and UniCL~\cite{yang2022unicl} to set up our training hyperparameters.  For optimization, we use AdamW~\cite{kingma2014adam} with a weight decay of 0.05 for all models. We set the learning rate to 0.0005 for locked-text-gated-image experiments, and 0.00005 for locked-text experiments.
We use the same set of data augmentation and regularization as in \cite{yang2022unicl}.
For experiments with 10M retrieved samples, the models are trained for 32 epochs with a batch size of 4096.  For experiments with fewer retrieved samples, the training epochs are adjusted accordingly so that they have a similar number of optimization steps.
For all training, we used a cosine learning rate schedule, with 5000 iterations warmup.

\subsection{Our Retrieval System}

\label{sec:retrieval_system}

We implement our retrieval system using FAISS~\cite{johnson2019faiss}.  We use its Hierarchical Navigable Small World (HNSW) approximate $k$-NN lookup~\cite{malkov2018hnsw} to balance performance and efficiency.  Product quantization is used to reduce the index size. We use \href{https://github.com/criteo/autofaiss}{Autofaiss} to select the optimal hyperparameters for the index, and build the index using FAISS index factory.  For LAION-400M, the selected configuration is: \texttt{OPQ256\_768,IVF131072\_HNSW32,PQ256x8}.  We build two separate indexing systems for T2I and T2T retrieval.  For T2I retrieval, CLIP image features are used for building the indexing system.  For T2T retrieval, CLIP text features are used.

We benchmark below the latency and the recall of the HNSW $k$-NN lookup on a server with 64 CPU cores.  As shown below, the indexing system is able to retrieve the relevant vectors accurately and efficiently.

\begin{table}[h]
    \centering
    \footnotesize
    \vspace{-3mm}
    \begin{tabular}{cccc}
        Latency & R@1 & R@10 & R@20 \\
        \shline
        0.57ms & 84.8 & 94.8 & 97.5 \\
    \end{tabular}
\end{table}

\subsection{Cost Estimation}

\label{sec:details_cost}

We provide the cost estimation for the \shortname{} pipeline.  It includes feature extraction of the retrieval pool, indexing for the retrieval system, querying the indexing system to retrieve relevant image-text pairs, and finally model customization.

\paragraph{Feature extraction.}  For feature extraction using CLIP ViT-B/32 checkpoint, it takes around 250 T4 GPU hours, which equates to roughly a day on a desktop with 4x RTX 3090 GPUs.

\paragraph{Build index.}  We build our index system on a cloud VM with 24 CPU cores.  Using the selected configuration in Sec.~\ref{sec:retrieval_system}, we can build the index system using FAISS within 20 hours.  Note that we use the CPU version of FAISS, and do not leverage GPU acceleration for building the index.

\paragraph{Querying index.} As shown in Sec.~\ref{sec:retrieval_system}, generating the retrieval set for model customization using the indexing system is very efficient.  Typically, it takes less than 10 minutes to generate 10M indices for our retrieval pool.

\paragraph{Model customization.} We train most of our models on a compute node with 16$\times$V100 GPUs.  For ViT-L checkpoints, we use 2-node distributed training, each node with 16$\times$V100.  It takes around 16/28/42 hours to train the B32/B16/L14 checkpoint, respectively, with either locked-text or locked-text-gated-image tuning strategy.

\paragraph{Remarks.}  Note that the feature extraction and building index system only needs to be done \emph{once}.  They are readily available for \emph{any} queries from any domains.  To make this line of research more accessible, our retrieved subsets for both ELEVATER
and ImageNet will also be made available, with an easy-touse toolkit to download the subsets without storing the whole dataset locally. Therefore, researchers can directly run experiments on model customization.

In conclusion, \shortname{} framework provides an accessible way to explore the large-scale web-crawled image-text dataset, and effectively and efficiently customize the models to the domain-of-interest.  

\end{document}